\documentclass[11pt]{article}
\usepackage[margin=1in]{geometry}
\usepackage[T1]{fontenc}
\usepackage{lmodern}
\usepackage{amsmath,amssymb}
\usepackage{booktabs}
\usepackage[numbers,sort&compress]{natbib}
\usepackage[hidelinks]{hyperref}
\usepackage{graphicx}
\usepackage{rotating}
\usepackage[font=footnotesize,skip=3pt]{caption}
\makeatletter
\renewcommand\paragraph{\@startsection{paragraph}{4}{\z@}%
  {1.6ex \@plus .4ex \@minus .2ex}{-1em}{\normalfont\normalsize\bfseries}}
\makeatother
\def\malhavtwotwo{1}
\newif\ifvtwotwo
\ifdefined\malhavtwotwo\vtwotwotrue\else\vtwotwofalse\fi
\newcommand{\vtwotwopending}[2]{%
  \par\smallskip\noindent\fbox{\parbox{\dimexpr\linewidth-2\fboxsep-2\fboxrule\relax}{%
  \footnotesize\textbf{PENDING --- not yet generated.} Expected file:
  \texttt{\detokenize{#1}}. #2 No value is reported until this file is written
  by its generator from a committed artifact.}}\smallskip\par}
\newcommand{\vtwotwotable}[2]{%
  \IfFileExists{#1.tex}{\resizebox{\ifdim\width>\linewidth\linewidth\else\width\fi}{!}{\input{#1}}}%
  {\vtwotwopending{#1.tex}{#2}}}
\newcommand{\vtwotwowidetable}[2]{%
  \IfFileExists{#1.tex}{\resizebox{\ifdim\width>\textheight\textheight\else\width\fi}{!}{\input{#1}}}%
  {\vtwotwopending{#1.tex}{#2}}}
\newcommand{\vtwotwofigure}[2]{%
  \IfFileExists{#1.pdf}{\includegraphics[width=\textwidth]{#1.pdf}}%
  {\vtwotwopending{#1.pdf}{#2}}}
\newcommand{\vtwotwotext}[2]{%
  \IfFileExists{#1.tex}{\input{#1}}{\vtwotwopending{#1.tex}{#2}}}
\ifvtwotwo\else\errmessage{malha.tex needs \string\malhavtwotwo: the abstract and
  conclusion depend on the v2.2 results}\fi
\newcommand{\vttHoneDelta}{+679}
\newcommand{\vttHoneCI}{[$-$95, +1{,}609]}
\newcommand{\vttHoneBCa}{[+8, +1{,}829]}
\newcommand{\vttHoneN}{60}
\newcommand{\vttHoneExpected}{60}
\newcommand{\vttHoneVerdict}{not resolved}
\newcommand{\vttHoneHybridCheaper}{9}
\newcommand{\vttHoneFloorCheaper}{45}
\newcommand{\vttHoneTied}{6}
\newcommand{\vttHoneNonTied}{54}
\newcommand{\vttHoneMedian}{$-$375}
\newcommand{\vttHoneExtraTP}{9}
\newcommand{\vttHoneExtraFP}{123}
\newcommand{\vttHoneFPperTP}{14}
\newcommand{\vttHoneFPmin}{1}
\newcommand{\vttHoneFPmax}{7}
\newcommand{\vttHoneGainMin}{1{,}034}
\newcommand{\vttHoneGainMax}{13{,}967}
\newcommand{\vttHoneBreakEven}{581}
\newcommand{\vttHoneWithoutTopThree}{+52}
\newcommand{\vttHoneHighFPDelta}{$-$1{,}883}
\newcommand{\vttHoneHighFPCI}{[$-$2{,}917, $-$725]}
\newcommand{\vttHoneGridSupported}{1}
\newcommand{\vttHoneGridOpposite}{2}
\newcommand{\vttHoneGridUnresolved}{3}

\newcommand{\vttDeskewHybridDelta}{$-$1{,}776}
\newcommand{\vttDeskewHybridCI}{[$-$3{,}728, $-$227]}

\newcommand{\vttDeskewHybridACheaper}{51}
\newcommand{\vttDeskewHybridBCheaper}{3}

\newcommand{\vttVlmFloorHybridDelta}{$-$3{,}153}
\newcommand{\vttVlmFloorHybridCI}{[$-$5{,}167, $-$1{,}513]}

\newcommand{\vttVlmFloorHybridACheaper}{54}
\newcommand{\vttVlmFloorHybridBCheaper}{0}

\newcommand{\vttNaiveVlmDelta}{$-$888}
\newcommand{\vttNaiveVlmCI}{[$-$3{,}896, +1{,}824]}
\newcommand{\vttNaiveVlmN}{60}
\newcommand{\vttNaiveVlmACheaper}{38}
\newcommand{\vttNaiveVlmBCheaper}{13}

\newcommand{\vttEngineVlmDelta}{+5{,}310}
\newcommand{\vttEngineVlmCI}{[+3{,}099, +7{,}766]}

\newcommand{\vttEngineVlmACheaper}{29}
\newcommand{\vttEngineVlmBCheaper}{28}

\newcommand{\vttHtwoDelta}{+6{,}777}
\newcommand{\vttHtwoCI}{[+4{,}155, +9{,}851]}

\newcommand{\vttHtwoACheaper}{0}
\newcommand{\vttHtwoBCheaper}{53}

\newcommand{\vttHthreeDelta}{$-$3}
\newcommand{\vttHthreeCI}{[$-$1{,}904, +2{,}051]}
\newcommand{\vttHthreeN}{60}

\newcommand{\vttHfourDelta}{+2{,}456}
\newcommand{\vttHfourCI}{[+919, +4{,}370]}

\newcommand{\vttHfourACheaper}{1}
\newcommand{\vttHfourBCheaper}{20}
\newcommand{\vttHfourTied}{39}
\newcommand{\vttVlmFloorTP}{121}

\newcommand{\vttVlmFloorInjected}{123}
\newcommand{\vttVlmFloorSameAsClean}{60}
\newcommand{\vttVlmFloorWorlds}{60}

\newcommand{\vttLegacyNaiveVlmDelta}{+5{,}084}
\newcommand{\vttLegacyNaiveVlmCI}{[+1{,}476, +9{,}531]}
\newcommand{\vttLegacyFixedVlmScanDelta}{+2{,}163}
\newcommand{\vttLegacyFixedVlmScanCI}{[+157, +5{,}465]}
\newcommand{\vttHfiveDelta}{+7{,}514}
\newcommand{\vttHfiveCI}{[+3{,}481, +11{,}841]}
\newcommand{\vttHfiveN}{30}
\newcommand{\vttHfiveExpected}{30}
\newcommand{\vttHfiveVerdict}{supported}
\newcommand{\vttHfiveAgentCheaper}{18}
\newcommand{\vttHfiveFloorCheaper}{12}

\newcommand{\vttHfiveMedian}{+3{,}135}
\newcommand{\vttHfiveFloorTP}{15}
\newcommand{\vttHfiveFloorFP}{2}
\newcommand{\vttHfiveAgentTP}{46}
\newcommand{\vttHfiveAgentFP}{47}
\newcommand{\vttHfiveInjected}{75}

\newcommand{\vttHfiveGridSupported}{5}
\newcommand{\vttHfiveGridCells}{6}

\newcommand{\vttHfiveHolm}{0.002}
\newcommand{\vttHfiveSignP}{0.36}
\newcommand{\vttHfiveCompDelta}{$-$6{,}452}
\newcommand{\vttHfiveCompCI}{[$-$9{,}470, $-$3{,}918]}
\newcommand{\vttHfiveCompN}{60}
\newcommand{\vttHfiveFloorEL}{15{,}426}
\newcommand{\vttHfiveAgentEL}{7{,}912}
\newcommand{\vttHfiveSurfaceClass}{\texttt{scan\_cutoff\_delivery}}
\newcommand{\vttHfiveSurfaceN}{15}
\newcommand{\vttHfiveSurfaceFloorTP}{0}
\newcommand{\vttHfiveSurfaceAgentTP}{15}
\newcommand{\vttHfiveSurfaceShare}{+8{,}441}
\newcommand{\vttHfiveSurfacePct}{112}
\newcommand{\vttHfiveSemTP}{20}
\newcommand{\vttHfiveSemN}{45}
\newcommand{\vttHfiveSemFloorTP}{0}
\newcommand{\vttHfiveSemShare}{+1{,}194}

\newcommand{\vttHfiveOtherShare}{$-$2{,}121}
\newcommand{\vttHfiveOtherN}{15}
\newcommand{\vttHfiveOtherFloorTP}{15}
\newcommand{\vttHfiveOtherAgentTP}{11}
\newcommand{\vttHfiveOtherAgentFP}{40}
\newcommand{\vttHfiveRcptTP}{14}
\newcommand{\vttHfiveRcptN}{15}
\newcommand{\vttHfiveRcptExposure}{162}
\newcommand{\vttHfiveDanfeTP}{4}
\newcommand{\vttHfiveDanfeN}{15}
\newcommand{\vttHfiveDanfeExposure}{3{,}484}
\newcommand{\vttHfiveMixedTP}{2}
\newcommand{\vttHfiveMixedN}{15}
\newcommand{\vttHfiveMixedExposure}{3{,}323}
\newcommand{\vttHfiveNoSurfDelta}{$-$927}
\newcommand{\vttHfiveNoSurfCI}{[$-$3{,}455, +1{,}190]}

\newcommand{\vttHfiveNoSurfAgentCheaper}{11}
\newcommand{\vttHfiveNoSurfFloorCheaper}{18}
\newcommand{\vttHfiveNoSurfTied}{1}
\newcommand{\vttHfiveSemOnlyDelta}{+1{,}194}
\newcommand{\vttHfiveSemOnlyCI}{[+298, +2{,}326]}

\newcommand{\vttHfiveWithDelta}{+14{,}985}
\newcommand{\vttHfiveWithCI}{[+8{,}999, +20{,}831]}
\newcommand{\vttHfiveWithN}{15}

\newcommand{\vttHfiveWithoutDelta}{+43}
\newcommand{\vttHfiveWithoutCI}{[$-$2{,}005, +2{,}180]}
\newcommand{\vttHfiveWithoutN}{15}

\newcommand{\vttHfiveWithoutFloorCheaper}{10}

\newcommand{\vttHfiveNoSurfHighFPDelta}{$-$2{,}802}
\newcommand{\vttHfiveNoSurfHighFPCI}{[$-$5{,}403, $-$489]}

\newcommand{\vttHfiveNoSurfGridUnresolved}{4}
\newcommand{\vttHfiveNoSurfGridOpposite}{2}
\newcommand{\vttHfiveCompSurfaceFloorTP}{15}
\newcommand{\vttHfiveCompSurfaceAgentTP}{13}
\newcommand{\vttHfiveCompSurfaceN}{15}
\newcommand{\vttHfiveTextureFP}{4}
\newcommand{\vttHfiveTextureFPAny}{13}

\newcommand{\vttHoneRetrySeed}{601}
\newcommand{\vttHoneNoRetryDelta}{+703}
\newcommand{\vttHoneNoRetryCI}{[$-$84, +1{,}650]}
\newcommand{\vttHoneNoRetryN}{59}
\newcommand{\vttHtwoRetrySeed}{621}
\newcommand{\vttHtwoNoRetryDelta}{+6{,}883}
\newcommand{\vttHtwoNoRetryCI}{[+4{,}269, +10{,}084]}
\newcommand{\vttHtwoNoRetryN}{59}

\newcommand{\vttHthreeNoRetryDelta}{$-$10}
\newcommand{\vttHthreeNoRetryCI}{[$-$1{,}960, +1{,}980]}

\title{Priced, not passed: what accounting agents actually lose\\in a SPED-native audit environment}
\author{Caio Theodoro\qquad Matheus Theodoro\\
\small Avenza Cloud (\texttt{avenza.cloud})\\
\small \href{https://orcid.org/0009-0005-7343-981X}{ORCID 0009-0005-7343-981X} (C.T.)\quad\href{https://orcid.org/0009-0001-4061-4117}{ORCID 0009-0001-4061-4117} (M.T.)}
\date{First version 25 November 2026; corrected draft 30 November 2026}

\begin{document}
\maketitle

\begin{abstract}
Agent evaluations score tasks as passed or failed, or grade answers against a rubric.
The economy prices errors in currency, and asymmetrically: a missed defect in a
Brazilian tax filing costs its statutory fine times the probability of audit, and a
false finding costs the hours spent chasing it. \textbf{malha} generates a synthetic
company's complete fiscal year --- NF-e XMLs, double-entry journal, trial balances,
rendered scanned source documents, and the statutory ECD and EFD-Contribui\c{c}\~oes
filings --- injects defects from a frozen taxonomy (15 classes in v1; 19 classes
including 4 physical-scan classes in v2), re-derives every downstream artifact so the
world stays consistent, and prices an agent's audit in the fines the law charges
(Lei 8.218/1991, art.~12). The oracle recomputes artifact evidence rather than trusting
agent output; injected defect metadata is used to target gate validation, not as a
model-visible answer.

Three studies are reported, and the evidence supports a narrower verdict than earlier
drafts. The first two use $n{=}30$ locked worlds ($k{=}3$ for the primary API cells,
$k{=}1$ for deterministic and local 8B cells; paired bootstrap over worlds). First,
standalone frontier agents in the v1 surface beat trivial floors yet fail to beat the
deterministic rules engine; those v1 rows are exploratory because the pre-correction
filesystem exposed \texttt{world.json}, which contained defect-class metadata. Audited
v2 \texttt{agent\_v2\_*} trajectories did not read that file. Second (v2 / v2.1, D033),
moving \texttt{malha\_rules} into a mandatory pre-pass, adding a read-only SQL index and
a fail-closed hash-chained Trust Ledger, and wiring real scan pixels into
\texttt{inspect\_scan} puts \texttt{agent\_v2\_vlm+glm-flash} ahead of the original
\texttt{malha\_rules} floor across all three noise regimes. A stronger deterministic
floor reverses the reading: \texttt{ocr\_rules\_book4}, four visible-artifact auditor
rules added to the OCR floor, prices at R\$108/R\$1{,}858/R\$4{,}937 across
clean/mild/noisy versus R\$5{,}699/R\$6{,}447/R\$5{,}986 for the best agent.

Third (v2.2, \texttt{prereg-v3}), the worlds were regenerated with a defect-independent
scan inventory (\vttHoneExpected{} worlds) and stateless transcription floors and a
rules-plus-agent hybrid were added. \textbf{v2.2 is exploratory}: its
oracle guard failed, so by its own registration no claim on the new pool is
confirmatory, although every registered cell was eventually run in full
(\vttHoneN{} of \vttHoneExpected{} worlds; \vttHfiveN{} of \vttHfiveExpected{} on
the subtle-wording slice). Within that limit it changes four things. (i) Two v2.1 scan claims made in earlier drafts did not replicate
and are retracted as general claims. On the legacy pool, whose scan inventory could cue
defect presence, pixel vision was ahead of the naive \texttt{ocr\_rules} floor on
\texttt{noisy\_scan} (\vttLegacyNaiveVlmDelta{} BRL \vttLegacyNaiveVlmCI{}) and ahead of
the same model without pixels on the scan-class slice (\vttLegacyFixedVlmScanDelta{} BRL
\vttLegacyFixedVlmScanCI{}); on the regenerated pool the same contrasts are
\vttNaiveVlmDelta{} \vttNaiveVlmCI{} and \vttHthreeDelta{} \vttHthreeCI{}. (ii) The
hybrid is cheaper than the strong floor on the mean (\vttHoneDelta{} BRL \vttHoneCI{},
\vttHoneVerdict{}) but dearer in \vttHoneFloorCheaper{} of \vttHoneNonTied{} non-tied
worlds: it adds \vttHoneExtraTP{} true and \vttHoneExtraFP{} false positives to the
rules' findings, and the mean favours it only while a false positive is priced below
about R\$\vttHoneBreakEven{}. (iii) The same model used once per page as a transcriber
feeding the deterministic rules is cheaper than that hybrid in
\vttVlmFloorHybridACheaper{} worlds and dearer in \vttVlmFloorHybridBCheaper{}.
(iv) On a 30-world slice whose defect wording was written after the rules' keyword
lists were frozen, with clean scans, the registered contrast favours the text agent
over the strong floor (\vttHfiveDelta{} BRL \vttHfiveCI{}; cheaper in
\vttHfiveAgentCheaper{} of \vttHfiveN{} worlds, \vttHfiveAgentFP{} false positives
against \vttHfiveFloorFP{}), while with the original wording the same contrast is
\vttHfiveCompDelta{} \vttHfiveCompCI{}. That margin is one class, and its rewording
is a date format, not a meaning: a delivery stamp printed as day, month name and
year, which the floor's stamp pattern does not parse (floor
\vttHfiveSurfaceFloorTP{} of \vttHfiveSurfaceN{}, agent \vttHfiveSurfaceAgentTP{}
of \vttHfiveSurfaceN{}; \vttHfiveSurfaceShare{} of the difference). On the three
semantically reworded classes the agent finds \vttHfiveSemTP{} of \vttHfiveSemN{}
defects, and without the date-format class the contrast is \vttHfiveNoSurfDelta{}
\vttHfiveNoSurfCI{} (post hoc; floor cheaper in \vttHfiveNoSurfFloorCheaper{}
worlds, agent in \vttHfiveNoSurfAgentCheaper{}). The slice therefore does not show
that the agent reads meaning the rules miss; a rule written against the new lexicon
recovers all of these defects. malha does \emph{not} show agents exceeding strong deterministic
automation; on these synthetic worlds scan noise breaks the reading step, and
repairing the reading step restores the rules.

Separately (8B trainability), current SFT/GRPO evidence is diagnostic only: corrected
ledger-shaped GRPO trains mechanically but its locked-test policy emits zero
model-origin findings in 30/30 episodes, exactly matches SFT priced loss, and fails the
pre-registered horizon-shape gate. Worlds, code, trajectories and the Gate-4 analysis
artifacts of the first two studies (2{,}329 scored episodes) are prepared for release
with the corrected submission package; the v2.2 artifacts are described in
\S\ref{sec:v22}.
\end{abstract}

\section{Introduction}
\label{sec:intro}

An auditor who misses R\$40{,}000 of unrecorded revenue and an auditor who flags 53
phantom defects are both wrong in a binary evaluation. They cost different money, and
the direction of the error matters: the miss is priced by the fine table times the
probability that the tax authority audits, and the phantom is priced in professional
hours. No mainstream agent benchmark prices either, and the deployment question for an
accounting agent is not whether it can pass an audit task but what it costs to let one
run. A firm choosing between a deterministic rules engine, an OCR regex pipeline, a
standalone LLM, or a deterministic-first hybrid VLM needs that answer in currency,
under declared enforcement and scan-noise regimes, with uncertainty attached.

\textbf{malha} prices the audit in the fines the law charges. The study has four
parts under three registrations (\texttt{PRE-REGISTRATION.md},
\texttt{PRE-REGISTRATION-V2.md} and \texttt{PRE-REGISTRATION-V3.md}; which of their
tags exist in the repository this revision is built from, and what a reader can
inspect, is stated in Appendix~\ref{app:registration}; every post-freeze amendment and harness audit
logged in \texttt{docs/DECISIONS.md}, D001--D058). No part is a confirmatory
result as it stands: the status of each is stated where it is reported.
\begin{enumerate}
\setlength{\itemsep}{2pt}\setlength{\parskip}{0pt}
\item \textbf{Why standalone agents lose (v1, \S\ref{sec:capability}--\ref{sec:taxonomy}).}
  On 15 text-based defect classes, frontier agents exhibit apparent substantive judgment
  (0.578 recall where rules are blind, now leak-contaminated/exploratory) yet fail to beat \texttt{malha\_rules} priced ---
  even with \texttt{run\_check} in-loop --- because pagination stamina and
  non-selective reliability leak free engine-detectable defects.
\item \textbf{How deterministic-first architecture narrows the gap on the legacy pool (v2 / v2.1,
  \S\ref{sec:envv2}--\ref{sec:results}).} Putting \texttt{malha\_rules} in a mandatory
  pre-pass, exposing a read-only SQLite index and fail-closed hash-chained Trust
  Ledger, adding 4 scanned-document defect classes across three physical noise regimes,
  and fixing three v2.0 harness defects uncovered by code audit (D033) separates hybrid
  agents ahead of the original \texttt{malha\_rules} floor (H1) across all noise and
  most enforcement regimes. On that pool pixel-level vision also overtook the naive
  \texttt{ocr\_rules} floor under scan noise; that reading did not replicate
  (item 3). The corrected stronger \texttt{ocr\_rules\_book4} floor remains ahead of
  every agent, so the result is gap-narrowing against the original floor, not
  deterministic-floor reversal.
\item \textbf{What a regenerated pool takes back (v2.2,
  \S\ref{sec:v22}; exploratory).} With the scan inventory made defect-independent, the two v2.1
  pixel-vision contrasts lose their sign, a rules-plus-agent hybrid is dearer than
  the strong floor in most worlds because of the false positives it adds, and the
  cheapest pipeline measured is the vision model used as a page transcriber in front
  of the same deterministic rules. On a slice whose wording the frozen keyword lists
  do not cover, the registered contrast favours the text agent
  (\vttHfiveDelta{} BRL \vttHfiveCI{} against the strong floor), but the margin is a
  date format the floor's stamp pattern does not parse; on the semantically reworded
  classes the agent finds \vttHfiveSemTP{} of \vttHfiveSemN{} defects and, without
  the date-format class, is not cheaper (\vttHfiveNoSurfDelta{}
  \vttHfiveNoSurfCI{}, post hoc). The oracle guard of this study failed, so every one of these
  readings is exploratory.
\item \textbf{Why 8B imitation and bounded local RL still fail (\S\ref{sec:trainability}).}
  Supervised fine-tuning on verified traces produces format-perfect policies that emit
  zero model-origin findings. Corrected ledger-shaped GRPO escapes the mechanical
  zero-variance wall but not the behavioural one: its locked-test policy still emits
  zero model-origin findings, exactly matches SFT priced loss, and fails the shape gate.
\end{enumerate}

\section{Related work}
\label{sec:related}

\paragraph{Agent benchmarks and cost-sensitive evaluation.} OSWorld \citep{osworld},
WorkArena \citep{workarena} and WebArena \citep{webarena} score task completion in
operating-system and web environments; $\tau$-bench \citep{taubench} scores
tool-agent-user dialogue; Terminal-Bench \citep{terminalbench} scores terminal tasks.
Scores are binary or rubric-shaped. GDPval \citep{gdpval} grades economically flavoured
deliverables against rubrics; the grade is points, not money, and the verifier is a
grader rather than a recomputation. While classical decision-theoretic learning
formalises asymmetric misclassification costs \citep{elkan2001}, mainstream agent
benchmarks have not adopted currency-priced statutory loss: a binary score has no axis
on which a missed R\$40{,}000 and a missed R\$400 differ.

\paragraph{Accounting NLP and bookkeeping benchmarks.} Document-level financial NLP
benchmarks such as FinQA \citep{finqa} and FiNER \citep{finer139} measure numerical
reasoning and XBRL entity tagging rather than multi-turn investigation. Recent
accounting-agent benchmarks --- FinBalance \citep{finbalance}, APEX-Accounting
\citep{apexaccounting}, and CAB \citep{cab} --- model journal entries and ledger
reconstruction, but target books rather than government-specified regulatory filings,
are eval-only, and do not price audit outcomes in statutory fines.

\paragraph{Reinforcement learning with verifiable rewards.} GRPO \citep{grpo} and the
RLVR lineage \citep{deepseekr1} train against programmatic reward. Recent analyses of
R1-style post-training highlight both the tendency of supervised fine-tuning to
memorise surface patterns \citep{sftmemorizes} and the pathology of zero-advantage
groups when rollout variance vanishes \citep{dapo,drgrpo}. Negative results on
long-horizon tool-use RLVR remain scarce in print; the failure mode measured here ---
SFT collapsing erratic base behaviour into a zero-variance fixed point --- is reported
in full because the mechanism, not the failure, is the contribution.

\section{The environment}
\label{sec:env}

\paragraph{World generation.} A world is a complete fiscal year of a synthetic
Brazilian \emph{lucro real} company: NF-e XML documents, the double-entry journal,
monthly trial balances, the ECD (Escritura\c{c}\~ao Cont\'abil Digital; 26 covered
records) and the EFD-Contribui\c{c}\~oes (33 covered records) --- the filing formats
companies submit to the Receita Federal \citep{sped}. Generation is seed-deterministic:
same seed, byte-identical world, enforced by test. CNPJs pass check-digit validation
and are drawn from the reserved synthetic \texttt{99.xxx.xxx} base convention
(Appendix~\ref{app:registration}).

\paragraph{Defect injection (v1 surface).} Each v1 world draws defects from a frozen
15-class taxonomy (Table~\ref{tab:taxonomy}, Appendix~\ref{app:spec}; v2 adds four
scan-substantive classes, \S\ref{sec:envv2}) split into two groups.
\emph{Engine-detectable} classes (11 classes across L1 layout, L2 accounting invariants,
L3 cross-filing rules, and L3-assisted sequence checks) --- duplicated postings,
control-total forgery, unbalanced entries, XML/filing inconsistencies --- are
irregularities a deterministic program can in principle always find. \emph{Substantive}
classes (4 classes: \texttt{cutoff\_violation}, \texttt{pis\_cofins\_ineligible\_credit},
\texttt{transitoria\_parking}, \texttt{expense\_personal\_mixed}) require judging
competence, tax-credit eligibility, account classification, or personal-spend substance.
Injection is followed by re-derivation: downstream artifacts (trial balances, J100/J150
statements, M-block apura\c{c}\~ao) are recomputed so the world stays internally
consistent with the defect present, except for two post-derivation classes that mutate
derived totals directly.

\paragraph{Tool surface and oracle by recomputation (v1 surface).} This paragraph
describes the v1 episode surface measured in \S\ref{sec:capability}--\ref{sec:taxonomy};
the v2 surface that replaces it is specified in \S\ref{sec:envv2}. Agents audit through a 6-tool
interface over a virtual filing filesystem with a 60-tool-call budget:
\texttt{list\_records}, \texttt{query\_records}, \texttt{read\_document},
\texttt{run\_check} (which executes the deterministic L1--L3 engine),
\texttt{flag\_finding}, and \texttt{submit}. The agent never sees \texttt{truth/} or
\texttt{defects.json}. Scoring is citation-anchored (D007): a finding counts as a true
positive only if \texttt{defect\_class} matches and at least one citation resolves to
the defect's anchor record and key; otherwise it is a false positive. Consequently, the
\texttt{flag\_everything} floor --- which emits one unanchored blanket finding per
class (15 findings) --- scores 0.0 recall and pays the full false-positive penalty.
The verifier recomputes every checkable quantity from the artifacts themselves; the
pilot gate required 100\% agreement between oracle verdicts and injected truth across
all 15 classes.

\paragraph{The loss model.} Expected loss is priced in cents-exact decimal BRL from
Lei 8.218/1991, art.~12 \citep{lei8218}:
\begin{equation}
\label{eq:loss}
\mathcal{L} = \sum_{d \notin \mathcal{M}} E(d; R, p) + c_{\text{fp}} \cdot n_{\text{fp}},
\end{equation}
where $\mathcal{M}$ is the set of matched defects, $R$ is annual gross revenue
(\texttt{receita\_bruta\_periodo}), $n_{\text{fp}}$ is the number of unmatched findings,
and $c_{\text{fp}} = \text{R\$}250.00$ is the professional rework cost per false
positive. Missed-defect exposure $E(d; R, p)$ follows the statutory basis of each class
(Table~\ref{tab:taxonomy}):
\begin{itemize}
\setlength{\itemsep}{1pt}\setlength{\parskip}{0pt}
\item \emph{Presentation} (art.~12, I): $0.005\,R$.
\item \emph{Operation} (art.~12, II): $\min(0.05\,v_d,\; 0.01\,R)$, where $v_d$ is the
  affected document or posting value.
\item \emph{Substantive}: $p \cdot (1 + 0.75)\,v_d$, pricing principal plus the 75\%
  \emph{multa de of\'icio} times $p = P(\text{autua\c{c}\~ao})$.
\item \emph{Operation + autua\c{c}\~ao} (\texttt{unrecorded\_nfe}; in v2 also
  \texttt{scan\_value\_alteration}): the sum of the art.~12, II penalty and the
  substantive term.
\end{itemize}
$P(\text{autua\c{c}\~ao})$ is a declared judgment parameter rather than an empirical
fact; three regimes are reported: conservative $0.05$, neutral $0.15$, and aggressive
$0.40$.

\begin{figure}[!htb]
\centering
\includegraphics[width=\textwidth]{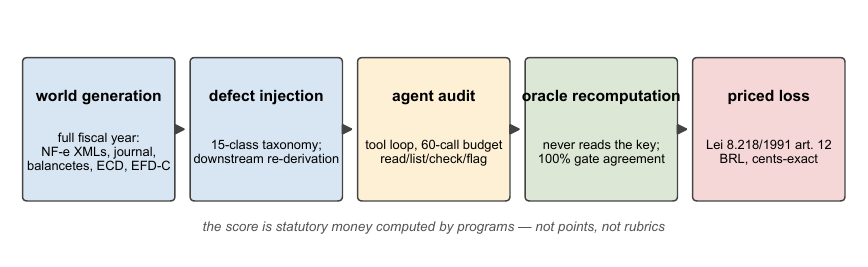}
\caption{The pipeline, drawn for the v1 surface (15-class taxonomy, 60-call budget).
A complete synthetic fiscal year is generated seed-deterministically; defects are
injected and downstream artifacts re-derived; the agent audits through a tool loop; the
oracle verifies by recomputation; the score is statutory money. The five stages are
unchanged in v2, which uses the 19-class taxonomy, a deterministic pre-pass outside the
agent budget, a SQL index, the Trust Ledger and a scan layer (\S\ref{sec:envv2}). ``Never
reads the key'' means the oracle never accepts agent output or the injected-defect list
as a verdict; \S\ref{sec:limitations} states where injected metadata still targets which
evidence is recomputed.}
\label{fig:pipeline}
\end{figure}

\section{Capability measurement}
\label{sec:capability}

\paragraph{Protocol.} Thirty worlds, seeds 201--230, each drawing one substantive and
one engine-detectable defect in mixed rotation across the 15 classes. Each arm runs
$k{=}3$ independent episodes per world at temperature $0.0$ under the 60-tool-call
budget with identical system prompt and tool definitions --- crucially,
\texttt{run\_check} is exposed to the agents throughout and noted in the prompt as
available for structural checks. The statistical unit is the world: episodes within a
world share its defects and are not independent evidence, so intervals are paired
bootstrap (10{,}000 resamples) over worlds. Headline numbers come from the committed
artifacts \texttt{rollout\_c2\_h2\_k3\_30.json} and
\texttt{rollout\_c2\_ds\_k3\_30\_merged.json} (see ``Disclosures and availability''). Arms:
\texttt{file\_as\_is} (submit nothing), \texttt{flag\_everything} (15 unanchored
findings, adding $15 \times 250 = 3{,}750.00$ BRL to \texttt{file\_as\_is}),
\texttt{malha\_rules} (scripted \texttt{run\_check} + flag every engine output), and
the frontier agent loops \texttt{gemini-hydrogen} and \texttt{deepseek-v4-pro}, with
smaller-sample evidence from \texttt{gemini-flash} ($n{=}5$) and \texttt{gemini-pro}
($n{=}4$, partial) in \S\ref{sec:taxonomy}.

\paragraph{Measurement status after D042.} These v1 rows are exploratory, not
confirmatory. The pre-correction v1 filesystem exposed \texttt{world.json}, which
contained defect-class metadata, and most audited legacy v1 trajectories read it. The
results below remain useful as a failure-mode and environment-design record, but they
cannot establish clean standalone-agent capability or a clean v1/v2 causal contrast.

\paragraph{The comparator.} \texttt{malha\_rules} is blind to substantive classes by
design choice, not by law of nature: every rule it lacks is codifiable in principle.
The honest question is therefore not whether an agent beats the engine in the abstract,
but what the engine's blindness is worth against what the agent costs. Both sides of
that trade are priced by $P(\text{autua\c{c}\~ao})$, so the comparison is reported as a
regime map rather than a single verdict.

\paragraph{The headline table.} Table~\ref{tab:headline} prices the arms under the
neutral profile. Both agents beat both floors decisively: hydrogen is $-$12{,}029.36
[$-$17{,}914.51, $-$6{,}785.94] against \texttt{file\_as\_is} and $-$15{,}779.36
[$-$21{,}664.51, $-$10{,}535.94] against \texttt{flag\_everything}; deepseek is
$-$10{,}405.61 [$-$15{,}196.99, $-$5{,}948.72] against \texttt{file\_as\_is} and
$-$14{,}155.61 [$-$18{,}946.99, $-$9{,}698.72] against \texttt{flag\_everything}. The
apparent capability is substantial but leak-contaminated. Against the deterministic engine, however, hydrogen's margin is
positive on the mean with the interval crossing zero (+3{,}191.33), and deepseek is
decisively worse (+4{,}815.08, interval excluding zero). Even with \texttt{run\_check}
inside their tool loop, neither standalone agent beats the incumbent. By contrast, a
post-hoc analytical union of \texttt{malha\_rules} with \texttt{gemini-hydrogen}'s
committed episodes --- retaining engine findings unconditionally and adding the agent's
zero-FP findings --- drops mean loss to 5{,}072.94 BRL/episode, separating cleanly
ahead of \texttt{malha\_rules} at $-$4{,}066.79 [$-$6{,}230.57, $-$2{,}257.20].

\begin{table}[!htb]
\centering\small
\begin{tabular}{lrr}
\toprule
arm & mean loss (BRL/ep) & margin vs \texttt{malha\_rules} [95\% CI] \\
\midrule
\texttt{flag\_everything} & 28{,}110.42 & +18{,}970.69 \\
\texttt{file\_as\_is} & 24{,}360.42 & +15{,}220.69 \\
\texttt{deepseek-v4-pro} & 13{,}954.81 & +4{,}815.08 [+1{,}333.23, +8{,}815.25] \\
\texttt{gemini-hydrogen} & 12{,}331.06 & +3{,}191.33 [$-$1{,}532.25, +8{,}326.65] \\
\texttt{malha\_rules} & 9{,}139.73 & --- \\
\midrule
\texttt{malha\_rules} $\cup$ \texttt{hydrogen}$^{\dagger}$ & 5{,}072.94 & $-$4{,}066.79 [$-$6{,}230.57, $-$2{,}257.20] \\
\bottomrule
\end{tabular}

\caption{Mean expected loss per episode, neutral enforcement profile, 30 worlds
$\times$ 3 episodes. Positive margin is worse than \texttt{malha\_rules}. Deepseek is
separated worse; hydrogen's interval crosses zero with a positive mean. $^{\dagger}$Post-hoc
analytical union of committed \texttt{malha\_rules} and \texttt{gemini-hydrogen}
episodes (since every hydrogen episode has zero FPs and either catches both defects or
exhausts budget with zero findings, the union loss is exact from
\texttt{rollout\_c2\_h2\_k3\_30.json}).}
\label{tab:headline}
\end{table}

\begin{figure}[!htb]
\centering
\includegraphics[width=0.62\textwidth]{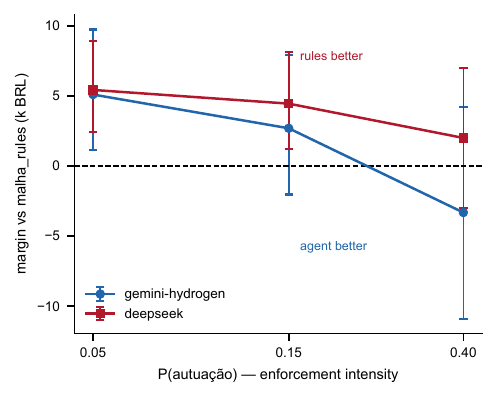}
\caption{The regime map. Paired margin against \texttt{malha\_rules} (thousand BRL) as
a function of P(autua\c{c}\~ao), log scale; negative is agent ahead. Bars are 95\%
paired bootstrap over worlds.}
\label{fig:regime}
\end{figure}

\paragraph{The regime map.} Table~\ref{tab:regime} and Figure~\ref{fig:regime} re-score
logged trajectories under all three enforcement regimes
(\texttt{loss\_profile\_sensitivity.json}). Re-scoring reconstructs findings from
JSONL tool-call logs rather than the live \texttt{AuditoriaEnv} result object; boundary
and re-run differences shift neutral point values slightly relative to
Table~\ref{tab:headline} ($-$509 BRL on hydrogen, $-$376 BRL on deepseek; D024), so
Table~\ref{tab:headline} carries the committed neutral point estimates while
Table~\ref{tab:regime} carries the regime-map shape. Under conservative enforcement both
agents are decisively worse, intervals separated: the agents' non-selective misses cost
money while the engine's blindness is cheap. As enforcement intensifies the engine's
structural blindness is increasingly priced, and hydrogen --- the arm with genuine
substantive recall (\S\ref{sec:slice}) --- crosses into the lead with the interval still
crossing zero. Deepseek never crosses. Which arm a deployment should choose depends on
how much enforcement the deployer believes in; the paper does not make that judgment
for the reader.

\begin{table}[!htb]
\centering\footnotesize
\begin{tabular}{lll}
\toprule
P(autua\c{c}\~ao) & hydrogen vs rules [95\% CI] & deepseek vs rules [95\% CI] \\
\midrule
0.05 (conservative) & +5{,}086.80 [+1{,}107.09, +9{,}729.39] & +5{,}417.32 [+2{,}442.37, +8{,}927.65] \\
0.15 (neutral) & +2{,}682.31 [$-$2{,}043.76, +7{,}901.51] & +4{,}439.02 [+1{,}218.89, +8{,}132.84] \\
0.40 (aggressive) & $-$3{,}328.92 [$-$10{,}909.89, +4{,}230.98] & +1{,}993.27 [$-$2{,}981.44, +6{,}965.46] \\
\bottomrule
\end{tabular}

\caption{Paired margin against \texttt{malha\_rules} as a function of enforcement
intensity, BRL per episode (re-scored from trajectories; D024). Under weak enforcement
both agents are decisively worse; under aggressive enforcement hydrogen's sign flips,
interval crossing zero.}
\label{tab:regime}
\end{table}

\section{The class-conditional slice}
\label{sec:slice}

If agents see what the engine cannot, why does the engine win the neutral regime?
Table~\ref{tab:slice} attributes recall and loss by defect kind across all 90 episodes
per arm (artifact \texttt{class\_slice.json}, alongside the exact \texttt{malha\_rules}
breakdown from \texttt{rollout\_c2\_h2\_k3\_30.json}).

\begin{table}[!htb]
\centering\small
\begin{tabular}{lrrrrr}
\toprule
arm & rec$_{\text{subs}}$ & rec$_{\text{eng}}$ & loss$_{\text{subs}}$ & loss$_{\text{eng}}$ & fp cost \\
\midrule
\texttt{gemini-hydrogen} & 0.578 & 0.578 & 4{,}722.84 & 7{,}099.20 & 0.00 \\
\texttt{deepseek-v4-pro} & 0.289 & 0.533 & 6{,}247.85 & 7{,}264.24 & 66.67 \\
\texttt{malha\_rules}$^{\dagger}$ & 0.000$^{*}$ & 0.933 & \multicolumn{2}{c}{9{,}048.06 (combined)} & 91.67 \\
\bottomrule
\end{tabular}

\caption{Class-conditional recall and loss attribution, 90 episodes per arm, BRL per
episode (agent rows from trajectory re-scoring \texttt{class\_slice.json};
$^{\dagger}$\texttt{malha\_rules} exact from \texttt{rollout\_c2\_h2\_k3\_30.json}:
28/30 engine-detectable worlds caught, 11 FPs across 30 worlds). $^{*}$Zero by
construction: the engine implements no rule for substantive classes.}
\label{tab:slice}
\end{table}

Three observations carry the measurement. First, the capability is real: hydrogen
recalls 57.8\% of substantive defects --- instances the engine is definitionally
incapable of catching. On the slice where the comparison favours the agent, the agent
delivers what the incumbent cannot. Second, the failure is non-selectivity rather than
absence of judgment: hydrogen's recall on engine-detectable defects is the same 0.578,
dropping --- at the exact rate it drops substantive defects --- irregularities that
\texttt{malha\_rules} catches on 93.3\% of worlds (28/30) at 91.67 BRL/episode FP cost.
Hydrogen loses more to uncaught engine-detectable defects (7{,}099 BRL per episode)
than to the substantive ones (4{,}723 BRL) that justify its use. Third, precision is a
model property rather than an agent property: hydrogen's false-positive cost across 90
episodes is exactly zero, whereas over-flagging belongs to specific models
(\S\ref{sec:taxonomy}), not to agentic auditing as such.

The structural implication is sharper than ``give the agent a rules tool.''
\texttt{run\_check} was \emph{already} exposed as a tool in the loop and named in the
system prompt, yet agents still exhausted their 60-call budget without emitting
findings on 38--42\% of episodes. Merely offering the engine voluntarily does not buy
its reliability. Running \texttt{malha\_rules} as a \emph{mandatory programmatic
pre-pass} --- outside the agent's budget, unioned with the agent's substantive findings
--- captures the $-$4{,}067 BRL/episode gain shown in Table~\ref{tab:headline} and is
the exact hybrid architecture the environment is built to test end to end next.

\section{The failure-mode taxonomy}
\label{sec:taxonomy}

\begin{figure}[!htb]
\centering
\includegraphics[width=\textwidth]{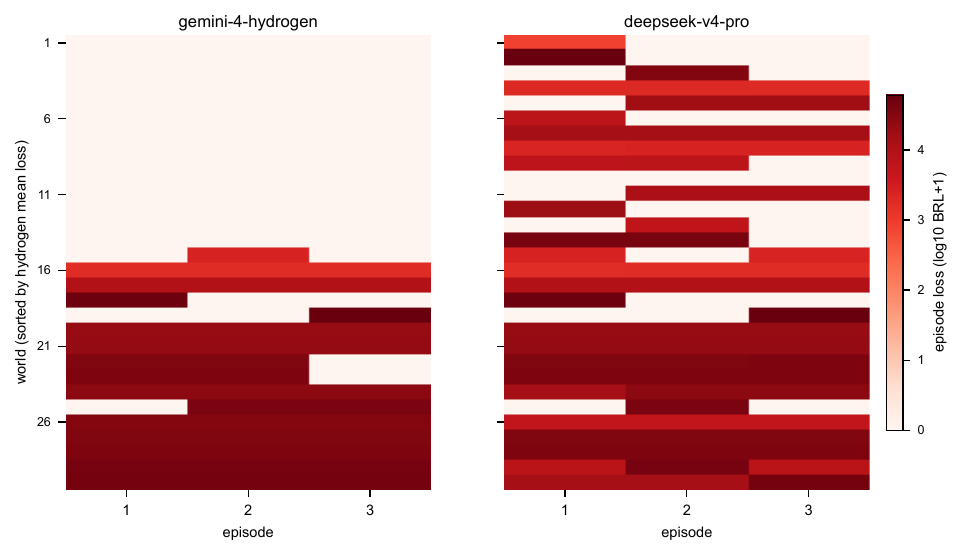}
\caption{Failure geometry: per-episode expected loss, log scale, 30 worlds $\times$ 3
episodes, rows sorted by hydrogen mean loss. Hydrogen's misses form solid horizontal
bands --- deterministic, identical across episodes. Deepseek's form ragged rows ---
stochastic engagement. Same task, same budget, different geometry.}
\label{fig:geometry}
\end{figure}

Per-episode loss geometry separates four failure families that binary scoring conflates
(Figure~\ref{fig:geometry}). \emph{gemini-flash} ($n{=}5$ worlds, seeds 101--105,
$k{=}3$, \texttt{rollout\_c2\_k3.json}; margin vs.\ \texttt{malha\_rules} +4{,}134.03
[$-$2{,}346.40, +12{,}220.37]) misses stochastically per episode: seed 103 scores
[74.3k, 0, 41.1k] across its three identical episodes, a coin flip per run.
\emph{gemini-pro} (partial $n{=}4$ worlds, seeds 101--104, $k{=}3$,
\texttt{rollout\_c2\_pro\_k3.json}, abandoned due to API rate limiting; margin
$-$2{,}781.03 [$-$9{,}945.01, +956.84]) over-flags on multi-finding worlds, with
per-world precision 0.30--0.56 (2.0--5.0 FPs/episode), the expensive direction of
error. \emph{gemini-hydrogen} ($n{=}30$) is blind deterministically per world/class:
14/30 worlds are completely clean across all three episodes and 10/30 are deterministic
total misses ([X, X, X] at 61 calls), leaving whole defect configurations systematically
invisible rather than stochastically dropped. \emph{deepseek-v4-pro} ($n{=}30$) engages
bimodally: careful investigation with high precision (0.67--1.0 across worlds, mean FP
cost 66.67 BRL) when engaged, stochastic total whiffs otherwise (seed 201 scores
[0, 0, 55.8k]), with graded partial credit between.

Whiff rates, flag-spam and class-blindness average into the same accuracy number. Only
the priced, per-episode, per-class view separates them, which is the methodological
claim of the paper: the geometry of failure is a model property, and it is invisible
without money on the axes.

\section{The v2 environment: evidence, noise, and commitment}
\label{sec:envv2}

The v1 measurement established a horizon problem: agents explore but do not
commit inside the budget. Version 2 rebuilds the episode surface around three
architectural interventions, each answering a v1 failure mode, and adds a
physical evidence layer that v1 lacked entirely.

\paragraph{The scan layer.} Every world now renders a deterministic sample of
its documents as scanned images: travel-expense receipts always, DANFEs,
NFS-e and delivery canhotos by sampling, plus any document carrying defect
evidence. Rendering is a deterministic pixel pipeline (vendored DejaVu fonts,
seeded rotation/noise) --- same seed, byte-identical PNG, enforced by test.
Three noise regimes structure the multimodal difficulty axis: \texttt{clean}
(straight render), \texttt{mild\_scan} (small rotation, light noise) and
\texttt{noisy\_scan} (larger rotation, heavier degradation). Four new
scan-substantive defect classes exist only in this layer: the books are
internally consistent and the irregularity is visible solely in the physical
document --- leisure items on a business-travel receipt
(\texttt{scan\_personal\_expense}), a January delivery stamp on a December
NF-e (\texttt{scan\_cutoff\_delivery}), a DANFE total above the mutually
consistent book and XML values (\texttt{scan\_value\_alteration}), and
scan-visible descriptions of non-creditable goods
(\texttt{scan\_ineligible\_credit\_desc}).

\paragraph{OCR as a cached, dual-fidelity artifact.} OCR runs once at
generation time; episodes read the cache, keeping rollouts fast and
byte-reproducible. Two caches exist per scan. The \emph{oracle cache}
(\texttt{truth/ocr\_perfect}) is a perfect layout extraction --- it is the
ground truth of what the document says, noise-independent, and the oracle
recomputes scan-defect presence from it. The \emph{agent cache} is
perfect-layout extraction under \texttt{clean}, but real Tesseract OCR over
degraded pixels under both noise regimes. The noise axis is therefore real:
mean extraction confidence falls from 0.96 (clean) to 0.69--0.77 (noisy), and
the deterministic OCR floor pays for it monotonically
(Section~\ref{sec:results}).

\paragraph{SQL analytical surface.} The v1 fixed-shape record queries are
replaced by a SQLite index over every SPED record type, the NF-e XML corpus,
the postings ledger and the OCR cache. Agents write arbitrary SQL. This moves
the bottleneck from schema navigation to judgment: the interesting question
is what the agent does with unlimited correct access to the data.

\paragraph{The trust ledger and fail-closed commitment.} Every tool
observation mints a hash-chained evidence receipt; every
\texttt{flag\_finding} must cite receipts whose payloads actually contain the
cited evidence, and the verifier rejects the finding otherwise --- rejected
findings are telemetry, never scorer false positives. The deterministic
pre-pass (the v1 engine) runs outside the agent's budget and its findings are
committed to the ledger before the episode starts; they price as the arm's
output whether or not the agent ever acts. The ablation is enforced
architecturally: \texttt{agent\_v2\_sql} shares everything with
\texttt{agent\_v2\_ledger} except the fail-closed gate. Absence evidence
required one verifier refinement (D028): a citation of a vacated sequence
number is backed by receipts observing its numeric neighbours, since the
missing record itself exists nowhere.

\paragraph{Horizon compression as a design constraint.} The v2 surface was
engineered to fit small-model training: fixed episode overhead compressed
from $\sim$3{,}050 to $\sim$900 tokens (D026), tool observations hard-capped,
and SQL output truncated aggressively. The constraint proved load-bearing in
an unexpected place: frontier-model audit episodes run 5{,}000--7{,}000 tokens
regardless of surface compression, because observation payloads dominate ---
which is why the 8B training data could not be curated from live agent traces
at the pre-registered 3{,}500-token gate and is instead synthesized by a
scripted expert playing the real environment under the same verification
rules (D028; Section~\ref{sec:trainability}).

\section{v2 results, D033 harness audit, and v2.1 separation}
\label{sec:results}

\paragraph{How to read this section after v2.2.} Everything in this section was measured
on the legacy locked pool (seeds 501--530), whose scan inventory was not
defect-independent (\S\ref{sec:limitations}). The numbers are kept as measured. Two
readings drawn from them did not replicate on the regenerated pool and are retracted
as general claims (\S\ref{sec:v22}, Table~\ref{tab:v22replication}): that pixel vision
is ahead of the naive \texttt{ocr\_rules} floor on \texttt{noisy\_scan}, and that pixel
access adds value over the same model reading OCR text. Where the text below says that
pixel access ``isolates'', ``doubles'' or ``separates'', read it as a description of
the legacy pool only. The comparison against the original \texttt{malha\_rules} floor
kept its sign on the regenerated pool.

\paragraph{The v2.0 matrix across the noise axis.} Table~\ref{tab:v2grid} and
Table~\ref{tab:v2contrasts} report the canonical Gate-4 rescoring of all
committed v2.0 and v2.1 episodes ($n=30$ locked worlds, seeds 501--530, $k=3$
for the primary API arms, neutral enforcement, R\$250 FP rework; generated directly from
\texttt{results/v2/analysis.json} by \texttt{scripts/build\_v2\_tables.py}). Error rows are
excluded before scoring rather than imputed. The pre-fix Gemini cell is incomplete:
177 noisy-scan rows are unusable (80 provider-quota errors and 97 harness
initialization errors), clean seed 512 is absent (29/30 worlds),
and noisy seeds 513--516 have only one or two successful episodes each. Those pre-fix
rows are retained as measurement history, not confirmatory v2.1 evidence; the two primary
v2.1 arms are complete 270/270 across noise.
The scan-noise axis is real: \texttt{ocr\_rules} degrades monotonically from
R\$6{,}241 (\texttt{clean}) to R\$7{,}991 (\texttt{mild\_scan}) to R\$11{,}070
(\texttt{noisy\_scan}), a +77.4\% loss increase as Tesseract garbles numeric
totals and item descriptions. Arms blind to scans (\texttt{file\_as\_is},
\texttt{flag\_everything}, \texttt{malha\_rules}) are invariant across noise by
construction.

In the pre-fix v2.0 matrix, both frontier flash arms priced numerically below
\texttt{malha\_rules} on \texttt{clean} (+4{,}635 BRL for
\texttt{gemini-flash}, +1{,}970 BRL for \texttt{deepseek-flash}), but neither
95\% paired bootstrap interval excluded zero (H1 not separated), and
\texttt{ocr\_rules} beat both text-only arms across every noise regime. On H2
(\texttt{agent\_v2\_sql} vs.\ \texttt{agent\_v2\_ledger}), neither flash arm is
an over-flagger, so the ledger's fail-closed gate yields no statistically
separated loss delta.

\paragraph{Where the loss lives (v2.0).} Loss decomposition on the completed
v2.0 cells locates the agent's deficit almost entirely in missed-defect fines,
not over-flagging: for \texttt{agent\_v2\_ledger+deepseek-flash} on clean
worlds, FP rework is 1.5\% of total priced loss and missed-defect fines are
98.5\% (0.9\% FP share for \texttt{gemini-flash}). The ledger mechanism
therefore cannot move the outcome for these non-over-flagging flash models ---
H2's premise is absent --- and the headline gap to the floor is a recall gap.
Per-class recall separates three distinct causes: (i) on \texttt{sequence\_gap},
0/12 recall was an instrumentation artifact of the D033
\texttt{adjudicate\_warning} schema bug rather than an agent reasoning failure;
(ii) on clean scans, \texttt{ocr\_rules} hits 100\% while text-only agents miss
a fraction of \texttt{scan\_value\_alteration} and
\texttt{scan\_ineligible\_credit\_desc}; and (iii) on book-side judgment
classes (\texttt{cutoff\_violation} 0/18, \texttt{transitoria\_parking} 0/9,
\texttt{pis\_cofins\_ineligible\_credit} 0--8\%), both the floor and the v2.0
flash arms score near zero (in contrast to v1's \texttt{gemini-hydrogen}
0.578 substantive recall, reflecting both the D030 flash-model downgrade and
v2.0's scan-focused system prompt). Unlocking the judgment slice and testing
true pixel-level vision under scan noise are the explicit targets of the D033
v2.1 protocol.

\paragraph{D033 harness audit and v2.1 separation.} A post-v2.0 code audit
(D033 in \texttt{docs/DECISIONS.md}) uncovered three implementation defects
confounding the v2.0 matrix:
\begin{enumerate}
\setlength{\itemsep}{1pt}\setlength{\parskip}{0pt}
\item \textbf{Broken \texttt{adjudicate\_warning} schema.} \texttt{TOOLS\_V2}
  advertised \texttt{(warning\_id, verdict: "confirmed"|"benign")}, while
  \texttt{AuditoriaEnvV2.\_t\_adjudicate} expected \texttt{(finding\_id,
  verdict: "confirm"|"dismiss")}. Across all v2.0 trajectories, 228 of 243
  adjudication calls failed with schema errors, artificially forcing 0/12
  recall on \texttt{sequence\_gap} for every agent arm (vs.\ 4/4 for
  \texttt{malha\_rules}).
\item \textbf{Mismatched SQL \texttt{schema\_text}.} \texttt{KEY\_COLS}
  advertised nonexistent column names across six tables (\texttt{dc},
  \texttt{descr}, \texttt{vl\_tot\_ap}, \texttt{nnf}, \texttt{vnf},
  \texttt{bbox}), burning agent turns on SQL error recovery.
\item \textbf{Scan-only prompt tunnel vision and text-only \texttt{inspect\_scan}.}
  \texttt{SYSTEM\_V2} listed a 4-bullet checklist only for \texttt{scan\_*}
  classes, starving book-substantive classes, and no v2.0 arm received scan PNG
  pixels.
\end{enumerate}

Fixing these defects and balancing \texttt{SYSTEM\_V2} (with frozen v2.0 rows
preserved untouched) transforms the outcome on the locked 30-world test pool.
Evaluated on identical seeds 501--530 ($k=3$, 180 episodes across
\texttt{clean} and \texttt{noisy\_scan}), \texttt{agent\_v2\_fixed+deepseek-flash}
cuts mean neutral expected loss on \texttt{clean} by 46.3\% (R\$13{,}749
$\to$ \textbf{R\$7{,}387}), \textbf{separating the exploratory v2.1 analogue of H1} against
\texttt{malha\_rules} at \textbf{+8{,}332 BRL [+4{,}116, +13{,}235]}. The mean
advantage remains positive across all six sensitivity cells
(conservative/neutral/aggressive $\times$ R\$250/R\$1{,}500); paired bootstrap
separation holds in five of six, with only conservative enforcement at
R\$1{,}500 FP cost crossing zero. Per-class recall unlocks across every book-side
judgment class where \texttt{ocr\_rules} is 0\% by construction:
\texttt{sequence\_gap} rises $0\% \to 92\%$ (11/12),
\texttt{cutoff\_violation} $0\% \to 28\%$ (5/18),
\texttt{transitoria\_parking} $0\% \to 33\%$ (3/9),
\texttt{pis\_cofins\_ineligible\_credit} $0\% \to 25\%$ (3/12), and
\texttt{expense\_personal\_mixed} $56\% \to 78\%$ (7/9).

\begin{figure}[!htb]
\centering
\includegraphics[width=\textwidth]{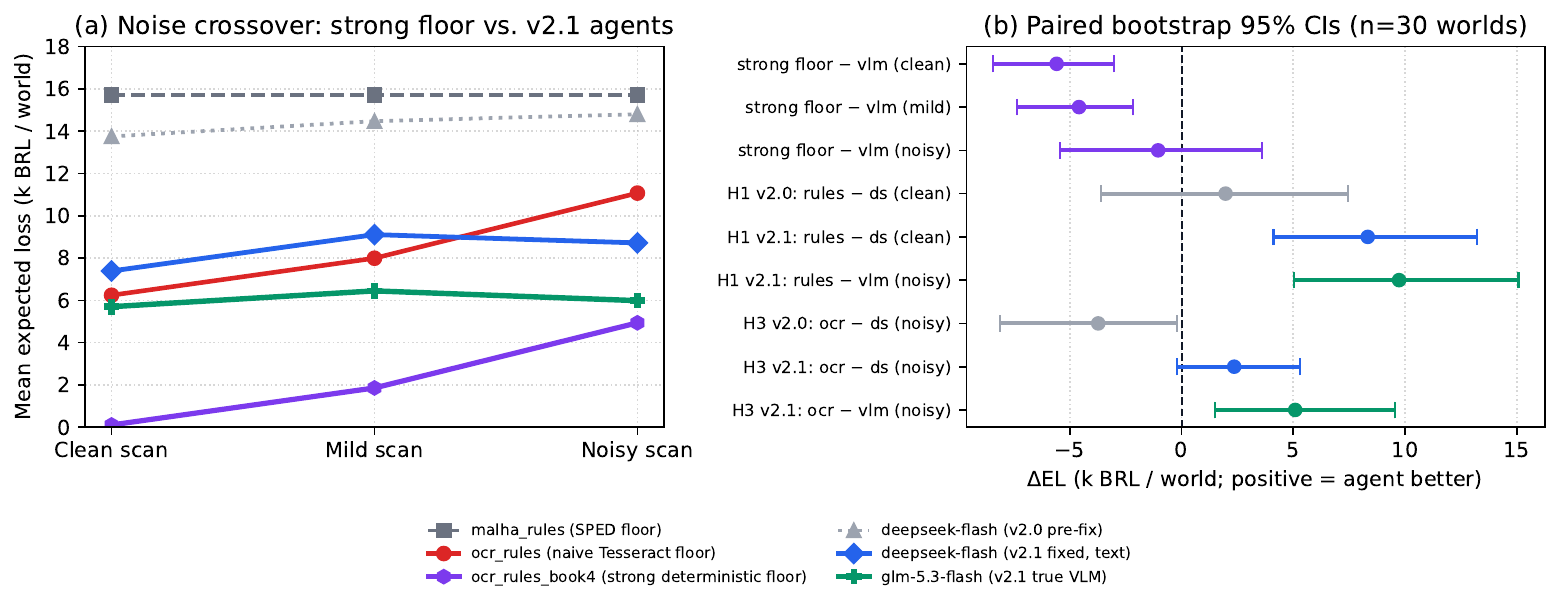}
\caption{v2.1 noise crossover and paired bootstrap contrasts (neutral profile,
R\$250 FP rework; generated from \texttt{results/v2/analysis.json}, 2{,}329
scored episodes; API cells are $n=30$ worlds $\times$ $k=3$ where complete, while
scripted and local 8B cells are deterministic/single-run $k=1$). (a)
\texttt{ocr\_rules} degrades +77.4\% from \texttt{clean} to \texttt{noisy\_scan},
while text-only \texttt{agent\_v2\_fixed+deepseek-flash} degrades +18.0\% and
true pixel-vision \texttt{agent\_v2\_vlm+glm-flash} degrades only +5.0\%.
The strong deterministic floor (purple) is lower-loss than every agent on clean and
mild scans and numerically lower on noisy scans. (b) Paired 95\%
bootstrap CIs: the post-hoc v2.1 arms separate against the original \texttt{malha\_rules}
floor, while the noisy-scan pixel advantage is decisive only against naive OCR. Registered
H1 on the v2.0 \texttt{agent\_v2\_ledger} arms did not separate; v2.1 contrasts are
exploratory amendments, not confirmatory H1 evidence. Legacy pool only: the noisy-scan
pixel advantage did not replicate on the regenerated pool
(Table~\ref{tab:v22replication}).}
\label{fig:v21crossover}
\end{figure}

The noise pattern supports a narrower noisy-scan result against naive OCR,
not the full pre-registered H3. Under \texttt{clean} scans,
\texttt{ocr\_rules} leads text-only \texttt{agent\_v2\_fixed+deepseek-flash}
non-significantly (R\$6{,}241 vs.\ R\$7{,}387; $-$1{,}146 [$-$3{,}980,
+1{,}501]); under \texttt{mild\_scan}, the two remain statistically tied
(R\$7{,}991 vs.\ R\$9{,}107; $-$1{,}116 [$-$5{,}406, +2{,}611]); under
\texttt{noisy\_scan}, \texttt{ocr\_rules} degrades +77.4\% to R\$11{,}070 while
text-only \texttt{agent\_v2\_fixed+deepseek-flash} degrades only +18.0\% to
R\$8{,}714 (+2{,}356 [$-$222, +5{,}293]). Enabling true pixel-level vision via
\texttt{inspect\_scan} (\texttt{agent\_v2\_vlm+glm-flash}, complete 270/270
episodes across all three noise levels on seeds 501--530) produces a monotonic
separation across the noise axis (Figure~\ref{fig:v21crossover}): paired against
\texttt{ocr\_rules}, $\Delta$EL progresses from +542 BRL [$-$2{,}492, +3{,}559]
on \texttt{clean} (R\$5{,}699 vs.\ R\$6{,}241) to +1{,}544 BRL [$-$1{,}167,
+4{,}580] on \texttt{mild\_scan} (R\$6{,}447 vs.\ R\$7{,}991) to
\textbf{+5{,}084 BRL [+1{,}476, +9{,}531]} on \texttt{noisy\_scan} (R\$5{,}986
vs.\ R\$11{,}070). This is decisive only for the noisy-scan endpoint in the overall-loss
comparison; the clean and mild endpoints do not exclude zero. Whereas
\texttt{ocr\_rules} degrades +77.4\% from \texttt{clean} to \texttt{noisy\_scan}
and text-only \texttt{agent\_v2\_fixed+deepseek-flash} degrades +18.0\%, true
pixel vision degrades only \textbf{+5.0\%} (R\$5{,}699 $\to$ R\$5{,}986).
Against \texttt{malha\_rules}, \texttt{agent\_v2\_vlm+glm-flash} separates the
exploratory v2.1 analogue of H1 at every noise level (\texttt{clean} \textbf{+10{,}020 [+5{,}106, +15{,}755]},
\texttt{mild\_scan} \textbf{+9{,}272 [+3{,}984, +15{,}498]},
\texttt{noisy\_scan} \textbf{+9{,}733 [+5{,}020, +15{,}080]}), and on
\texttt{noisy\_scan} it beats \texttt{ocr\_rules} by \textbf{+2{,}532 BRL [+9,
+6{,}150]} even when restricted strictly to the four scan-substantive classes
(95\% paired bootstrap CI excluding zero). H3 as pre-registered is therefore
\textbf{not confirmed}: the scan-slice mild-scan contrast is $-$243 BRL
[$-$1{,}199, +722], and the noisy-scan positive is thin --- the four scan classes have
only 3--6 instanced worlds each in the locked pool (5 for \texttt{scan\_cutoff\_delivery},
3 for \texttt{scan\_ineligible\_credit\_desc}, 3 for \texttt{scan\_personal\_expense},
and 6 for \texttt{scan\_value\_alteration}), and exact-citation strict rescoring makes the noisy scan-slice interval
cross zero (D044). Pixel vision
directly recovers OCR-garbled fields where Tesseract regex collapses: on
\texttt{noisy\_scan}, a same-model controlled contrast holding \texttt{glm-5.3-flash}
weights, prompt, SQL index, and Trust Ledger identical (\texttt{agent\_v2\_fixed+glm-flash},
text-only OCR, R\$7{,}733/world, vs.\ \texttt{agent\_v2\_vlm+glm-flash}, pixel vision,
R\$5{,}986/world; 90/90 episodes each) shows that pixel access alone \emph{doubles}
recall on both OCR-degraded scan classes --- \texttt{scan\_value\_alteration} rises
from 28\% (5/18) to \textbf{56\%} (10/18) and \texttt{scan\_ineligible\_credit\_desc}
from 22\% (2/9) to \textbf{44\%} (4/9) --- separating decisively on the four-class
scan slice at \textbf{+2{,}163 BRL [+157, +5{,}465]} (+1{,}748 BRL [$-$783, +5{,}294]
overall), alongside 100\% on \texttt{sequence\_gap} (12/12) and 44\% on
\texttt{cutoff\_violation} (8/18).

\paragraph{The stronger deterministic floor changes the headline.}
A correction-pass floor, \texttt{ocr\_rules\_book4}, adds four auditor rules over
agent-visible artifacts: personal-expense narration markers, December transitory-account
parking, document-versus-recognition-month cutoff, and PIS/COFINS credit claimed on
non-creditable input CFOPs. These rules never read \texttt{defects.json} or truth
artifacts. On the locked clean pool they recover every world-level instance of the four
book-side judgment classes with zero class-level false-positive worlds. Priced across
clean/mild/noisy scans, the combined floor means are R\$108/R\$1{,}858/R\$4{,}937,
versus R\$5{,}699/R\$6{,}447/R\$5{,}986 for \texttt{agent\_v2\_vlm+glm-flash}.
The paired floor-minus-agent contrasts are $-$5{,}591 BRL [$-$8{,}453, $-$3{,}025] on
clean and $-$4{,}589 BRL [$-$7{,}360, $-$2{,}165] on mild scans: the deterministic floor
is decisively lower-loss. On noisy scans the floor remains numerically ahead by
1{,}048 BRL but the interval crosses zero [$-$5{,}437, +3{,}582]. Consequently H1/H3 are
retained only as comparisons against the originally registered \texttt{malha\_rules}
and \texttt{ocr\_rules} floors, not as evidence that agents exceed strong deterministic
automation.

\paragraph{Scoring-exploit sensitivity.} The primary scorer permits multiple citations
on a finding because a real audit finding may legitimately cite complementary records.
That design is exploitable: a zero-knowledge control that emits one finding per taxonomy
class with the union of visible record/key candidates recovers 66/71 clean-pool defect
instances in the uncapped scorer (mean recall 0.917, mean loss R\$9{,}119), but collapses
to 1/71 under one- or three-citation caps in the committed control artifact
(\texttt{results/v2/citation\_stuffing.json}). This control is a scorer diagnostic, not
an audit capability result. The measured agents do not depend on that exploit in the
headline comparison: \texttt{agent\_v2\_vlm+glm-flash} averages 2.32 citations per
finding on noisy scans; cap-3 and cap-1 rescoring move its mean loss only from R\$5{,}986
to R\$6{,}083/R\$6{,}348, while \texttt{ocr\_rules\_book4} remains unchanged at
R\$4{,}937 (\texttt{analysis.json}).

\paragraph{The legacy v1 surface is pricedly inadequate under the same model.}
The post-D042 leak-free \texttt{agent\_v1\_postd042+glm-flash} contrast completed
90/90 clean locked episodes through the same \texttt{glm-5.3-flash} endpoint. Without
access to the pre-D042 \texttt{world.json} defect metadata, the legacy fixed-query,
pre-ledger interface prices at R\$35{,}155/world versus R\$5{,}699 for
\texttt{agent\_v2\_vlm+glm-flash}; the paired v1-minus-v2 delta is
\textbf{+R\$29{,}456 [+22{,}907, +36{,}019]} in v2.1's favour. The leak-free v1
surface is also decisively worse than \texttt{malha\_rules}: the rules-minus-v1
contrast is $-$19{,}437 BRL [$-$26{,}257, $-$12{,}840]. The pre-D042 v1 rows
(R\$23{,}987/world; v1-minus-v2 +R\$18{,}287 [+12{,}020, +24{,}947]) remain in the
tables only as leak-exposed measurement history. This is still an arm-level surface
comparison, not a decomposition of SQL, ledger, and pixel effects individually:
v1 differs from v2.1 in all three.

\begin{table}[!htb]
\centering\small
\begin{tabular}{lrrr}
\toprule
arm & clean & mild\_scan & noisy\_scan \\
\midrule
\texttt{flag\_everything} & 43,836 & 43,836 & 43,836 \\
\texttt{file\_as\_is} & 39,086 & 39,086 & 39,086 \\
\texttt{qwen3-8b-v2-sft} & 19,645 & --- & --- \\
\texttt{qwen3-8b-base} & 17,787 & --- & --- \\
\texttt{qwen3-8b-v2-grpo} & 19,645 & --- & --- \\
\texttt{malha\_rules} & 15,719 & 15,719 & 15,719 \\
\texttt{agent\_v2\_sql+deepseek-flash} & 13,103 & --- & --- \\
\texttt{agent\_v2\_ledger+deepseek-flash} & 13,749 & 14,470 & 14,799 \\
\texttt{agent\_v2\_sql+gemini-flash} & 12,070 & --- & --- \\
\texttt{agent\_v2\_ledger+gemini-flash} & 11,609 & 13,149 & 13,078 \\
\texttt{agent\_v2\_fixed+deepseek-flash} & 7,387 & 9,107 & 8,714 \\
\texttt{agent\_v2\_fixed+glm-flash} & 9,675 & 9,567 & 7,733 \\
\texttt{agent\_v2\_vlm+glm-flash} & 5,699 & 6,447 & 5,986 \\
\texttt{ocr\_rules} & 6,241 & 7,991 & 11,070 \\
\texttt{agent\_v1+glm-flash} & 23,987 & --- & --- \\
\texttt{agent\_v1\_postd042+glm-flash} & 35,155 & --- & --- \\
\texttt{malha\_rules\_book4} & 9,586 & 9,586 & 9,586 \\
\texttt{ocr\_rules\_book4} & 108 & 1,858 & 4,937 \\
\bottomrule
\end{tabular}

\caption{v2.0 and v2.1 (D033) mean neutral-profile expected loss (BRL per
world, R\$250 FP rework, lower is better) on the locked test pool (seeds
501--530, complete $n=30$ worlds $\times$ $k=3$ = 270/270 episodes each for
\texttt{fixed+deepseek-flash} and \texttt{vlm+glm-flash} across all three noise
levels, plus 90/90 episodes per noise level for the same-model text-only ablation
\texttt{fixed+glm-flash}, 90/90 for the post-D042 leak-free
\texttt{agent\_v1\_postd042+glm-flash} clean-surface contrast, and 90/90 pre-D042
legacy v1 rows retained only as leak-exposed measurement history). Row legend:
\texttt{agent\_v1+glm-flash} is the pre-D042 leak-exposed row and
\texttt{agent\_v1\_postd042+glm-flash} its leak-free rerun; \texttt{malha\_rules\_book4}
and \texttt{ocr\_rules\_book4} are the post-hoc strong floors (the four book-side rules
added to \texttt{malha\_rules} and to \texttt{ocr\_rules} respectively); scripted floors
and the three \texttt{qwen3-8b} rows are $k{=}1$; ``---'' marks a cell that was not run.
Rows appear in generator order, not sorted by loss. Auto-generated from
\texttt{results/v2/analysis.json} (2{,}329 scored episodes).}
\label{tab:v2grid}
\end{table}

\begin{table}[!htb]
\centering\footnotesize
\resizebox{\textwidth}{!}{
\begin{tabular}{llr}
\toprule
contrast (positive = second arm better) & $\Delta$EL [95\% CI] (BRL) & $n$ \\
\midrule
Registered H1 (v2.0): \texttt{malha\_rules} $-$ \texttt{ledger+gemini-flash} (clean) & +4,635 [-2,408, +11,862] & 29 \\
Registered H1 (v2.0): \texttt{malha\_rules} $-$ \texttt{ledger+deepseek-flash} (clean) & +1,970 [-3,586, +7,461] & 30 \\
Exploratory H1 analogue (v2.1): \texttt{malha\_rules} $-$ \texttt{fixed+deepseek-flash} (clean) & \textbf{+8,332 [+4,116, +13,235]} & 30 \\
Exploratory H1 analogue (v2.1 VLM): \texttt{malha\_rules} $-$ \texttt{vlm+glm-flash} (clean) & \textbf{+10,020 [+5,106, +15,755]} & 30 \\
Exploratory H1 analogue (v2.1 VLM): \texttt{malha\_rules} $-$ \texttt{vlm+glm-flash} (noisy) & \textbf{+9,733 [+5,020, +15,080]} & 30 \\
H2: \texttt{sql} $-$ \texttt{ledger} (gemini-flash, clean) & +875 [-1,419, +4,319] & 29 \\
H2: \texttt{sql} $-$ \texttt{ledger} (deepseek-flash, clean) & -646 [-2,536, +785] & 30 \\
Registered H3 (v2.0): \texttt{ocr\_rules} $-$ \texttt{ledger+gemini-flash} (clean) & \textbf{-5,170 [-9,545, -1,590]} & 29 \\
Exploratory H3 analogue (v2.1): \texttt{ocr\_rules} $-$ \texttt{fixed+deepseek-flash} (clean) & -1,146 [-3,980, +1,501] & 30 \\
Leak-free v1 surface: \texttt{agent\_v1\_postd042} $-$ \texttt{vlm+glm-flash} (clean) & \textbf{+29,456 [+22,907, +36,019]} & 30 \\
Strong floor: \texttt{ocr\_rules\_book4} $-$ \texttt{vlm+glm-flash} (clean) & \textbf{-5,591 [-8,453, -3,025]} & 30 \\
Strong floor: \texttt{ocr\_rules\_book4} $-$ \texttt{vlm+glm-flash} (mild) & \textbf{-4,589 [-7,360, -2,165]} & 30 \\
Strong floor: \texttt{ocr\_rules\_book4} $-$ \texttt{vlm+glm-flash} (noisy) & -1,048 [-5,437, +3,582] & 30 \\
Exploratory H3 analogue (v2.1 VLM): \texttt{ocr\_rules} $-$ \texttt{vlm+glm-flash} (clean) & +542 [-2,492, +3,559] & 30 \\
Exploratory H3 analogue (v2.1 VLM): \texttt{ocr\_rules} $-$ \texttt{vlm+glm-flash} (mild) & +1,544 [-1,167, +4,580] & 30 \\
Registered H3 (v2.0): \texttt{ocr\_rules} $-$ \texttt{ledger+gemini-flash} (noisy) & -2,008 [-6,755, +2,159] & 30 \\
Registered H3 (v2.0): \texttt{ocr\_rules} $-$ \texttt{ledger+deepseek-flash} (noisy) & \textbf{-3,729 [-8,127, -203]} & 30 \\
Exploratory H3 analogue (v2.1): \texttt{ocr\_rules} $-$ \texttt{fixed+deepseek-flash} (noisy) & +2,356 [-222, +5,293] & 30 \\
Exploratory H3 analogue (v2.1 VLM): \texttt{ocr\_rules} $-$ \texttt{vlm+glm-flash} (noisy) & \textbf{+5,084 [+1,476, +9,531]} & 30 \\
Vision ablation: \texttt{fixed+glm-flash} $-$ \texttt{vlm+glm-flash} (noisy) & +1,748 [-783, +5,294] & 30 \\
\bottomrule
\end{tabular}
}
\caption{v2.0 and v2.1 paired bootstrap contrasts over shared worlds (10{,}000
resamples, seed 11, neutral profile, R\$250 FP rework). Positive $\Delta$EL
favours the second arm; bold entries exclude zero.}
\label{tab:v2contrasts}
\end{table}

\begin{table}[!htb]
\centering\footnotesize
\resizebox{\textwidth}{!}{
\begin{tabular}{lrrrrrr}
\toprule
sensitivity cell & \texttt{fixed+ds} clean & \texttt{fixed+ds} mild & \texttt{fixed+ds} noisy & \texttt{vlm+glm} clean & \texttt{vlm+glm} mild & \texttt{vlm+glm} noisy \\
\midrule
conservative, R\$250 & \textbf{+2,686} & \textbf{+1,737} & \textbf{+2,189} & \textbf{+3,432} & \textbf{+2,788} & \textbf{+3,318} \\
conservative, R\$1500 & +1,755 & +585 & +1,216 & +1,793 & +1,011 & +1,485 \\
neutral, R\$250 & \textbf{+8,332} & \textbf{+6,612} & \textbf{+7,005} & \textbf{+10,020} & \textbf{+9,272} & \textbf{+9,733} \\
neutral, R\$1500 & \textbf{+7,401} & \textbf{+5,459} & \textbf{+6,032} & \textbf{+8,381} & \textbf{+7,494} & \textbf{+7,900} \\
aggressive, R\$250 & \textbf{+22,446} & \textbf{+18,797} & \textbf{+19,044} & \textbf{+26,490} & \textbf{+25,479} & \textbf{+25,770} \\
aggressive, R\$1500 & \textbf{+21,515} & \textbf{+17,644} & \textbf{+18,072} & \textbf{+24,851} & \textbf{+23,702} & \textbf{+23,937} \\
\bottomrule
\end{tabular}
}
\caption{H1 sensitivity grid: mean paired \texttt{malha\_rules}-minus-agent
expected-loss deltas (BRL/world) for the two complete v2.1 arms across enforcement
probability ($0.05/0.15/0.40$) and false-positive rework cost (R\$250/R\$1{,}500).
Positive values favour the agent; bold entries are paired-bootstrap separated
(10{,}000 resamples over worlds). The conservative/R\$1{,}500 row remains positive
on mean but is the only row whose intervals cross zero.}
\label{tab:v2sensitivity}
\end{table}

\paragraph{Scan-class noise decomposition.} Table~\ref{tab:v2scan} decomposes
v2.0/v2.1 recall across the four scan-substantive classes. Under \texttt{clean}
scans, 20 lines of regex (\texttt{ocr\_rules}) achieve 100\% recall because
the synthetic trigger phrases are lexically clean. Under \texttt{noisy\_scan},
Tesseract degradation collapses \texttt{ocr\_rules} to 0\% on
\texttt{scan\_ineligible\_credit\_desc} and 33\% on
\texttt{scan\_value\_alteration}; pixel vision is the only arm that improves
both classes in the same-model comparison on this pool. On the regenerated pool the
pixel agent reads those two classes about as poorly as the text agent
(Table~\ref{tab:v22scan}).

\begin{table}[!htb]
\centering\footnotesize
\begin{tabular}{llrrrrr}
\toprule
noise & scan class & \texttt{ocr} & \texttt{v2.0 ledger+ds} & \texttt{fixed+ds} & \texttt{fixed+glm} & \texttt{vlm+glm} \\
\midrule
\texttt{clean} & \texttt{scan\_cutoff\_delivery} & 100\% & 93\% & 87\% & 80\% & 100\% \\
 & \texttt{scan\_ineligible\_credit\_desc} & 100\% & 33\% & 56\% & 56\% & 89\% \\
 & \texttt{scan\_personal\_expense} & 100\% & 89\% & 100\% & 100\% & 100\% \\
 & \texttt{scan\_value\_alteration} & 100\% & 56\% & 56\% & 56\% & 61\% \\
\midrule
\texttt{mild\_scan} & \texttt{scan\_cutoff\_delivery} & 100\% & 87\% & 100\% & 80\% & 100\% \\
 & \texttt{scan\_ineligible\_credit\_desc} & 0\% & 22\% & 22\% & 0\% & 33\% \\
 & \texttt{scan\_personal\_expense} & 100\% & 89\% & 100\% & 100\% & 67\% \\
 & \texttt{scan\_value\_alteration} & 83\% & 44\% & 17\% & 61\% & 61\% \\
\midrule
\texttt{noisy\_scan} & \texttt{scan\_cutoff\_delivery} & 100\% & 100\% & 100\% & 100\% & 100\% \\
 & \texttt{scan\_ineligible\_credit\_desc} & 0\% & 33\% & 33\% & 22\% & 44\% \\
 & \texttt{scan\_personal\_expense} & 100\% & 100\% & 100\% & 67\% & 78\% \\
 & \texttt{scan\_value\_alteration} & 33\% & 28\% & 39\% & 28\% & 56\% \\
\bottomrule
\end{tabular}

\caption{Per-class recall on the four scan-substantive classes across noise
regimes for the deterministic OCR floor, the pre-fix v2.0 ledger arm, the
post-fix text-only arms, and the v2.1 pixel-vision arm. \texttt{ocr\_rules}
dominates \texttt{clean} scans (100\%) but collapses on noisy
\texttt{scan\_ineligible\_credit\_desc} (0\%) and
\texttt{scan\_value\_alteration} (33\%). On \texttt{noisy\_scan}, the
same-model GLM comparison isolates the pixel effect: 22\% $\to$ 44\% and
28\% $\to$ 56\% on those two classes.}
\label{tab:v2scan}
\end{table}

\begin{table}[!htb]
\centering\footnotesize
\resizebox{\textwidth}{!}{
\begin{tabular}{llrrrrr}
\toprule
arm & noise & episodes & prompt Mtok & output Mtok & total USD & mean USD/ep \\
\midrule
\texttt{agent\_v1\_postd042+glm-flash} & \texttt{clean} & 90/90 & 90.98 & 2.66 & 14.98 & 0.166 \\
\texttt{fixed+ds} & \texttt{clean} & 90/90 & 17.93 & 1.07 & 2.67 & 0.030 \\
\texttt{fixed+ds} & \texttt{mild\_scan} & 90/90 & 19.09 & 1.08 & 2.81 & 0.031 \\
\texttt{fixed+ds} & \texttt{noisy\_scan} & 90/90 & 17.49 & 1.02 & 2.59 & 0.029 \\
\texttt{fixed+glm} & \texttt{clean} & 90/90 & 15.59 & 0.30 & 2.49 & 0.028 \\
\texttt{fixed+glm} & \texttt{mild\_scan} & 90/90 & 17.06 & 0.30 & 2.71 & 0.030 \\
\texttt{fixed+glm} & \texttt{noisy\_scan} & 90/90 & 17.33 & 2.18 & 3.69 & 0.041 \\
\texttt{ledger+ds} & \texttt{clean} & 90/90 & 16.71 & 1.00 & 2.49 & 0.028 \\
\texttt{ledger+ds} & \texttt{mild\_scan} & 90/90 & 18.51 & 1.06 & 2.73 & 0.030 \\
\texttt{ledger+ds} & \texttt{noisy\_scan} & 90/90 & 17.89 & 1.13 & 2.69 & 0.030 \\
\texttt{ledger+gem} & \texttt{mild\_scan} & 90/90 & 29.56 & 0.15 & 22.73 & 0.253 \\
\texttt{ledger+gem} & \texttt{noisy\_scan} & 83/83 & 27.23 & 0.14 & 20.94 & 0.252 \\
\texttt{sql+ds} & \texttt{clean} & 90/90 & 16.56 & 1.04 & 2.49 & 0.028 \\
\texttt{vlm+glm} & \texttt{clean} & 90/90 & 14.57 & 2.03 & 3.20 & 0.036 \\
\texttt{vlm+glm} & \texttt{mild\_scan} & 90/90 & 17.19 & 2.08 & 3.62 & 0.040 \\
\texttt{vlm+glm} & \texttt{noisy\_scan} & 90/90 & 17.20 & 2.21 & 3.68 & 0.041 \\
\bottomrule
\end{tabular}
}
\caption{Measured API-token cost rollup by arm and noise, computed from captured
per-episode prompt/completion token counts in \texttt{results/v2/analysis.json}.
Dollar values use list prices in force during the runs (deepseek-flash
\$0.12/\$0.48 per million input/output tokens under the OpenCode discount,
glm-5.3-flash \$0.15/\$0.50, gemini-3.8-flash \$0.75/\$3.75). They exclude
cache-hit discounts, free-tier quota, subscription amortization, and grounding;
episodes without usage capture are visible in the episodes column rather than
estimated.}
\label{tab:v2costs}
\end{table}

\ifvtwotwo
\clearpage 
\section{v2.2: an exploratory rerun on a regenerated pool}
\label{sec:v22}

\noindent\fbox{\parbox{\dimexpr\linewidth-2\fboxsep-2\fboxrule\relax}{\footnotesize
\textbf{Status of this section.} The v2.2 registration (\texttt{prereg-v3}) was frozen
before the regenerated pools existed. Two things then went wrong and are reported as
such. The oracle guard failed on the pool registered as confirmatory, so by the
registration's own rule every claim on that pool is exploratory, and rerunning cannot
repair that: it needs a new pool under a new registration. The provider's usage limit
then interrupted the agent runs; they were resumed in a later usage window and every
registered cell, including the two for H5, now has one successful episode per world.
Two worlds entered after a retry that was not a provider error (world
\vttHtwoRetrySeed{} of the pixel agent, where the model emitted an invalid action, and
world \vttHoneRetrySeed{} of the hybrid, where the runner environment lacked a module
before any model call); each affected contrast is reported with and without that
world. Nothing in this section is a confirmatory result. Every
table, figure and verdict sentence below is generated from the artifacts in
\texttt{results/v22/}; a cell that was not run says so. Decisions D054--D057 in
\texttt{docs/DECISIONS.md} record the handling; D056 closes the status of this pool
as exploratory and opens no new registration, and D057 records the completed
measurement and what changed with it. The tables name the \vttHoneExpected{}-world pool
``confirmatory'' because that is its registered name; the prose calls it the
regenerated pool.}}

\paragraph{What v2.2 changes, and what it leaves alone.} Section~\ref{sec:limitations}
lists four restrictions on the v2.1 scan and floor results that are properties of the
legacy locked pool rather than of any agent: the scan inventory is not
defect-independent, the pool predates the D046 renderer correction, the v2.1 contrasts
were designed after inspecting the same pool, and the scan-class slice is thin. v2.2
regenerates the worlds under a registration frozen before the pool existed. The
design and gate tables below report which of those conditions were met. The 19-class
taxonomy, the loss model (Equation~\ref{eq:loss}), the scorer and the paired-world
bootstrap are unchanged. The legacy-pool rows in \S\ref{sec:results} are not rewritten
or pooled with v2.2 rows (D049, D051). Agent arms run one episode per world on one
flash-class model with the v2.1 prompt.

\paragraph{Design and registration.} The registered pools, seed ranges, world counts,
episodes per world, arms, hypotheses and decision rules are reproduced from the frozen
registration and the pool manifests rather than restated here
(Table~\ref{tab:v22design}).

\begin{table}[!htb]
\centering\footnotesize
\vtwotwotable{tables/v22_design}{One row per regime (fresh confirmatory pool; subtle
stress slice; legacy locked pool, new arms only): registration tag and commit, seed
range, worlds, episodes per world, \texttt{gen\_version}, arms run, and status
(confirmatory or exploratory).}
\caption{v2.2 design, generated from the frozen registration and the pool manifests.
``Confirmatory pool'' is the registered name of the regenerated pool; its status
column gives what it is as measured.}
\label{tab:v22design}
\end{table}

\paragraph{Gates: the inventory gate passed, the oracle gate did not.} The v2.1 scan
result is exploratory because scan count rose with the number of scan-class defects
(\texttt{results/v2/scan\_inventory\_shape.json}). On the regenerated pool every world
has the same number of scans and the same document-type mix whatever its defects
(Table~\ref{tab:v22inventory}), and each scan class is instanced in the same number of
worlds. That confound is removed. The oracle gate is a separate guard and failed: the
oracle recomputes \vttVlmFloorTP{} of the \vttVlmFloorInjected{} injected defects, the
two it misses sitting on aggregate statement rows its invariant does not check
(D054). No deterministic arm can find those two defects, and dropping their worlds
post hoc changes no registered decision (generated verdicts below), but the guard is
binary and its consequence is not negotiable after the fact.

\begin{table}[!htb]
\centering\footnotesize
\vtwotwotable{tables/v22_scan_inventory}{Visible scan PNG count and document-type mix by
number of scan-class defects per world, the instanced-world count per scan class, and
the oracle-gate agreement on the regenerated pool.}
\caption{Scan-inventory shape, class strata and oracle gate on the regenerated pool.}
\label{tab:v22inventory}
\end{table}

\paragraph{Degraded scans on the regenerated pool.} The question the legacy
pool could not settle is whether any agent arm is lower-loss than the maximised
deterministic floor once scans degrade, and whether a model used as a stateless
transcriber feeding deterministic rules does better or worse than the same model in an
open tool loop. Table~\ref{tab:v22grid} prices every arm; Table~\ref{tab:v22contrasts}
gives the registered paired contrasts; Figure~\ref{fig:v22ladder} orders the
architectures. Agent cells exist under \texttt{noisy\_scan} for all three agent arms
and under \texttt{clean} for the text agent only. The verdict on each
registered hypothesis is generated with the tables and printed next; the paragraphs
after it say what those verdicts do and do not mean.

\begin{table}[!htb]
\centering\small
\vtwotwotable{tables/v22_primary_grid}{Mean neutral-profile expected loss per world, arm
by scan-noise regime, with episodes completed per cell.}
\caption{v2.2 primary grid on the regenerated pool (BRL per world, lower is better).}
\label{tab:v22grid}
\end{table}

\begin{table}[!htb]
\centering\footnotesize
\vtwotwotable{tables/v22_contrasts}{Registered paired contrasts with 95\% paired-bootstrap
intervals and world counts, sign convention stated in the header row, confirmatory and
exploratory rows labelled.}
\caption{v2.2 paired contrasts over shared worlds. All rows are exploratory.}
\label{tab:v22contrasts}
\end{table}

\begin{figure}[!htb]
\centering
\vtwotwofigure{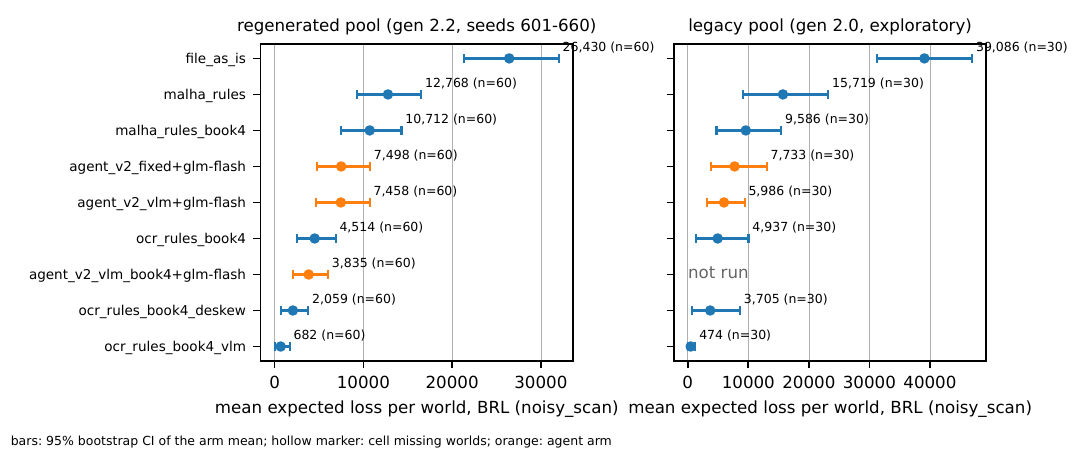}{Architectural ladder: mean expected loss with
95\% paired-bootstrap intervals for each rung, on the regenerated pool, with the legacy
pool shown separately and labelled.}
\caption{Mean expected loss by architecture under scan noise (exploratory). Means hide the per-world split reported in Table~\ref{tab:v22perworld}.}
\label{fig:v22ladder}
\end{figure}

\paragraph{Registered hypotheses, as measured.}
Each reading below is generated from \texttt{results/v22/registered\_analysis.json}. $\Delta(A,B)=\mathrm{EL}(A)-\mathrm{EL}(B)$ in BRL per world; the interval is the registered 95\% percentile paired bootstrap.
\begin{itemize}\itemsep2pt
\item \textbf{H1} (\texttt{ocr\_rules\_book4} vs.\ \texttt{agent\_v2\_vlm\_book4+glm-flash}; v22, noisy\_scan): \textbf{not resolved}, reported as \textbf{exploratory}. $\Delta=$\,$+679$ [$-95$, $+1{,}609$] over 60 of 60 worlds (lost: 0 for the first arm, 0 for the second; limit 3). Under \texttt{strict\_cap3}: $+679$ [$-95$, $+1{,}609$], same decision. Fragile: the BCa interval [$+8$, $+1{,}829$] gives a different decision. Per world: the second arm is cheaper in 9, the first in 45, tied in 6 (exact sign test $p=$\,$7\times10^{-7}$). The mean favours the second arm and the count of worlds favours the first: a few large worlds carry the mean. The decision holds in 3 of 6 cells of the enforcement $\times$ rework grid (of the other 3: 1 exclude zero with the second arm cheaper, 2 exclude zero with the first arm cheaper), so it is a cell claim. Post hoc, without the oracle-failed worlds (635, 647): $+466$ [$-229$, $+1{,}308$], $n=58$, same decision. World 601 of \texttt{agent\_v2\_vlm\_book4+glm-flash} is a retry after a failed first attempt (a missing Python module in the runner environment, before any model call); without it (post hoc): $+703$ [$-84$, $+1{,}650$], $n=59$, same decision. Why exploratory: guard 6.1 failed (results/v22/design/oracle\_gate\_confirmatory.json).
\item \textbf{H2} (\texttt{agent\_v2\_vlm+glm-flash} vs.\ \texttt{ocr\_rules\_book4\_vlm}; v22, noisy\_scan): \textbf{supported}, reported as \textbf{exploratory}. $\Delta=$\,$+6{,}777$ [$+4{,}155$, $+9{,}851$] over 60 of 60 worlds (lost: 0 for the first arm, 0 for the second; limit 3). Under \texttt{strict\_cap3}: $+6{,}826$ [$+4{,}213$, $+9{,}917$], same decision. Holm-adjusted permutation $p=0.0004$ (family of 4, 4 with data). Per world: the second arm is cheaper in 53, the first in 0, tied in 7 (exact sign test $p=$\,$2\times10^{-16}$). The decision is the same in every cell of the enforcement $\times$ rework grid. Post hoc, without the oracle-failed worlds (635, 647): $+6{,}748$ [$+4{,}036$, $+9{,}852$], $n=58$, same decision. World 621 of \texttt{agent\_v2\_vlm+glm-flash} is a retry after a failed first attempt (the model emitted an invalid action); without it (post hoc): $+6{,}883$ [$+4{,}269$, $+10{,}084$], $n=59$, same decision. Why exploratory: guard 6.1 failed (results/v22/design/oracle\_gate\_confirmatory.json).
\item \textbf{H3} (\texttt{agent\_v2\_fixed+glm-flash} vs.\ \texttt{agent\_v2\_vlm+glm-flash}; v22, noisy\_scan, scan-class slice): \textbf{not resolved}, reported as \textbf{exploratory}. $\Delta=$\,$-3$ [$-1{,}904$, $+2{,}051$] over 60 of 60 worlds (lost: 0 for the first arm, 0 for the second; limit 3). Under \texttt{strict\_cap3}: $-23$ [$-1{,}930$, $+2{,}009$], same decision. Holm-adjusted permutation $p=0.9984$ (family of 4, 4 with data). Fragile: the sign changes when one world is left out. Per world: the second arm is cheaper in 20, the first in 21, tied in 19 (exact sign test $p=$\,1.000). The decision is the same in every cell of the enforcement $\times$ rework grid. Post hoc, without the oracle-failed worlds (635, 647): $-3$ [$-1{,}933$, $+2{,}065$], $n=58$, same decision. World 621 of \texttt{agent\_v2\_vlm+glm-flash} is a retry after a failed first attempt (the model emitted an invalid action); without it (post hoc): $-10$ [$-1{,}960$, $+1{,}980$], $n=59$, same decision. Why exploratory: guard 6.1 failed (results/v22/design/oracle\_gate\_confirmatory.json).
\item \textbf{H4} (\texttt{ocr\_rules\_book4\_reocr} vs.\ \texttt{ocr\_rules\_book4\_deskew}; v22, noisy\_scan): \textbf{supported}, reported as \textbf{exploratory}. $\Delta=$\,$+2{,}456$ [$+919$, $+4{,}370$] over 60 of 60 worlds (lost: 0 for the first arm, 0 for the second; limit 3). Under \texttt{strict\_cap3}: $+2{,}456$ [$+919$, $+4{,}370$], same decision. Holm-adjusted permutation $p=0.0004$ (family of 4, 4 with data). Per world: the second arm is cheaper in 20, the first in 1, tied in 39 (exact sign test $p=$\,$2\times10^{-5}$). The decision is the same in every cell of the enforcement $\times$ rework grid. Post hoc, without the oracle-failed worlds (635, 647): $+2{,}540$ [$+984$, $+4{,}566$], $n=58$, same decision. Why exploratory: guard 6.1 failed (results/v22/design/oracle\_gate\_confirmatory.json).
\item \textbf{H5} (\texttt{ocr\_rules\_book4} vs.\ \texttt{agent\_v2\_fixed+glm-flash}; v22\_subtle, clean): \textbf{supported}, reported as \textbf{exploratory}. $\Delta=$\,$+7{,}514$ [$+3{,}481$, $+11{,}841$] over 30 of 30 worlds (lost: 0 for the first arm, 0 for the second; limit 2). Under \texttt{strict\_cap3}: $+7{,}143$ [$+3{,}076$, $+11{,}486$], same decision. Holm-adjusted permutation $p=0.0020$ (family of 4, 4 with data). Per world: the second arm is cheaper in 18, the first in 12, tied in 0 (exact sign test $p=$\,0.362). The decision holds in 5 of 6 cells of the enforcement $\times$ rework grid (of the other 1: 1 include zero), so it is a cell claim. Companion condition (point estimate $\le$ 0 on v22, clean): $-6{,}452$ [$-9{,}470$, $-3{,}918$] over 60 of 60 worlds, holds. By class (registration \S9.1; Table~\ref{tab:v22h5classes}): \texttt{scan\_cutoff\_delivery}, whose rewording is a date format, accounts for $+8{,}441$ of the difference and the three semantic classes for $+1{,}194$, with 20 of 45 semantic defects found by the second arm. Post hoc, without that class: $-927$ [$-3{,}455$, $+1{,}190$], second arm cheaper in 11 worlds, first in 18, tied in 1. Why exploratory: guard 6.1 failed (results/v22/design/oracle\_gate\_confirmatory.json).
\end{itemize}

\begin{table}[!htb]
\centering\footnotesize
\vtwotwotable{tables/v22_perworld}{Per-world split of the registered and post hoc
contrasts under noisy scans: worlds in which each arm is cheaper, and the hybrid's
added true and false positives.}
\caption{The same contrasts counted per world (regenerated pool, \texttt{noisy\_scan},
neutral profile, R\$250 per false positive). Positive $\Delta$ means $B$ is cheaper on
the mean; the count columns say in how many worlds. Generated by
\texttt{paper/build\_v22\_prose.py} with the registered scorer and bootstrap.}
\label{tab:v22perworld}
\end{table}

\paragraph{H1 per world: a favourable mean made of a few catches and many false alarms.}
The registered H1 reading is ``\vttHoneVerdict{}'': floor minus hybrid is
\vttHoneDelta{} BRL per world \vttHoneCI{} over \vttHoneN{} of \vttHoneExpected{}
worlds, and the reading is fragile (the BCa interval is \vttHoneBCa{}). The mean is
the wrong summary of what happened. Per world the hybrid is cheaper than its own
rules floor in \vttHoneHybridCheaper{} worlds, dearer in \vttHoneFloorCheaper{} and
tied in \vttHoneTied{}; the median difference is \vttHoneMedian{} BRL
(Table~\ref{tab:v22perworld}). In every shared world the hybrid has at least the floor's
true positives and at least its false positives. Across those worlds it added \vttHoneExtraTP{} true positives and
\vttHoneExtraFP{} false positives, about \vttHoneFPperTP{} false alarms per extra
catch. In each of the \vttHoneFloorCheaper{} worlds where it is dearer it caught
nothing the rules had missed and added between \vttHoneFPmin{} and \vttHoneFPmax{}
false positives. In each of the \vttHoneHybridCheaper{} worlds where it is cheaper it
caught at least one defect the rules had missed, removing between R\$\vttHoneGainMin{}
and R\$\vttHoneGainMax{} of exposure. Without the three largest differences the mean
is \vttHoneWithoutTopThree{}.

\paragraph{H1 is a statement about the price of a false positive.} Whether
\vttHoneExtraTP{} catches outweigh \vttHoneExtraFP{} false alarms depends on what a
false alarm costs, and that price is a declared assumption, not a measurement. The
mean favours the hybrid only while a false positive is priced below about
R\$\vttHoneBreakEven{}. At the registered primary price of R\$250 it does; at the
registered alternative of R\$1{,}500 the same contrast is \vttHoneHighFPDelta{}
\vttHoneHighFPCI{} under the neutral profile, in the floor's favour with the interval
excluding zero. Across the six cells of the registered enforcement $\times$ rework
grid the direction favours the hybrid with the interval excluding zero in
\vttHoneGridSupported{}, favours the floor with the interval excluding zero in
\vttHoneGridOpposite{}, and is unresolved in \vttHoneGridUnresolved{}. Where the
generated H1 sentence says the decision ``holds'' in some cells of the grid, it counts
cells with the same reading as the primary cell; it does not mean the hybrid is ahead
in them. Two further
cautions apply. The agent prompt, kept unchanged from v2.1, tells the model that a
rejected finding costs nothing, so the false-positive rate may be a property of that
prompt. And the scorer counts as false any finding where nothing was injected; these
findings were not audited by hand, so some may be reasonable flags on unrealistic
synthetic texture.

\paragraph{The registered floor was not the strongest floor.} H1 compares the hybrid
with \texttt{ocr\_rules\_book4} reading raw Tesseract text. Two pipelines without an
agent loop, both arms of the same registration, are lower-loss than the hybrid (post
hoc contrasts,
Table~\ref{tab:v22perworld}; every floor against every agent arm is in
Table~\ref{tab:v22floormatrix}). Deskewing the image before OCR and applying the same
rules is cheaper than the hybrid in \vttDeskewHybridACheaper{} worlds and dearer in
\vttDeskewHybridBCheaper{} (\vttDeskewHybridDelta{} \vttDeskewHybridCI{}). Using the
vision model once per page as a transcriber and applying the same rules is cheaper in
\vttVlmFloorHybridACheaper{} and dearer in \vttVlmFloorHybridBCheaper{}
(\vttVlmFloorHybridDelta{} \vttVlmFloorHybridCI{}). So the registered H1 reading, whichever way it had come out, would not
show an agent beating deterministic automation; it would show an agent against a
floor that the same registration expected to be surpassed. The hybrid cell is now
complete (\vttHoneN{} of \vttHoneExpected{} worlds) and the reading stayed
\vttHoneVerdict{}; without the retried world \vttHoneRetrySeed{} it is
\vttHoneNoRetryDelta{} \vttHoneNoRetryCI{} over \vttHoneNoRetryN{} worlds.

\paragraph{H2 and H4: what holds is about reading the page.} The two supported
readings concern transcription, not auditing. H4: undoing the synthetic rotation and
noise before OCR lowers the rules' loss (\vttHfourDelta{} \vttHfourCI{}; deskew
cheaper in \vttHfourBCheaper{} worlds, dearer in \vttHfourACheaper{}, no difference in
\vttHfourTied{}). This is a noisy-scan cell result: the same recipe raises loss on
clean images (Table~\ref{tab:v22grid}), it was selected on development seeds, and
the distortion it undoes is the generator's own transform. H2:
the vision model used as a transcriber in front of the rules is cheaper than the same
model as an agent (\vttHtwoDelta{} \vttHtwoCI{}; transcriber cheaper in
\vttHtwoBCheaper{} worlds, agent in \vttHtwoACheaper{}). With the model's transcript
the rules find \vttVlmFloorTP{} of \vttVlmFloorInjected{} defects, the same as on
perfect text, and price identically to the clean-scan floor in
\vttVlmFloorSameAsClean{} of \vttVlmFloorWorlds{} worlds. H2 is not a controlled test
of ``loop versus no loop'', as the registration disclosed before the freeze: the
agent sees scans downscaled, pays for each look out of a 25-call budget, and its
pre-pass lacks the four book-side rules the floor has. The arm that would separate
these, an agent on top of the transcript-fed rules, was not run. The ceiling is also
partly circular: the rules and the defect injector share a keyword list, so
``correct text restores the rules'' is in part the generator agreeing with itself.

\paragraph{What did not replicate from v2.1.} The registration wrote this branch
before the pool existed: if the noisy-scan advantage of pixel vision over the
text-only agent disappears under the fixed inventory, it is reported as a retraction
of the exploratory v2.1 reading, and a regenerated contrast that changes sign makes
the regenerated rows primary and the legacy rows historical. Both conditions are met
on the data available (Table~\ref{tab:v22replication}). The same-model pixel contrast
on the scan-class slice, \vttLegacyFixedVlmScanDelta{} \vttLegacyFixedVlmScanCI{} on
the legacy pool, is \vttHthreeDelta{} \vttHthreeCI{} over \vttHthreeN{} of
\vttHoneExpected{} worlds, and per class the pixel agent reads altered values and
ineligible descriptions about as poorly as the text agent (Table~\ref{tab:v22scan}).
The pixel agent against the naive \texttt{ocr\_rules} floor,
\vttLegacyNaiveVlmDelta{} \vttLegacyNaiveVlmCI{} on the legacy pool, is
\vttNaiveVlmDelta{} \vttNaiveVlmCI{} over \vttNaiveVlmN{} worlds, with the naive floor
cheaper in \vttNaiveVlmACheaper{} and the agent in \vttNaiveVlmBCheaper{}. We
therefore retract the v2.1 readings that pixel access isolates a scan-reading
advantage and that pixel vision is ahead of naive OCR under scan noise. The wording
matters: both regenerated intervals include zero, so
this is a failure to replicate with a changed sign, not a demonstrated reversal. The
legacy numbers stay in \S\ref{sec:results} as measurements on a pool whose inventory
could cue defects. One v2.1 reading kept its sign: both agents remain cheaper on the
mean than the original \texttt{malha\_rules} engine (\vttEngineVlmDelta{}
\vttEngineVlmCI{} for the pixel agent), although per world that is an even split
(\vttEngineVlmBCheaper{} against \vttEngineVlmACheaper{}).

\begin{table}[!htb]
\centering\footnotesize
\vtwotwotable{tables/v22_replication}{v2.1 legacy-pool contrasts beside the same
contrasts on the regenerated pool.}
\caption{v2.1 contrasts re-tested on the regenerated pool (\texttt{noisy\_scan},
neutral profile). Legacy values are from \texttt{results/v2/analysis.json}; regenerated
values use the registered scorer and bootstrap and, except the H3 row, are post hoc.
The pools differ in size, episodes per world and scan inventory, so the columns are a
replication check, not a paired comparison.}
\label{tab:v22replication}
\end{table}

\paragraph{The new floors on the legacy locked pool.} The new floors were also run on
the legacy locked pool (seeds 501--530) so that they can be read against
Table~\ref{tab:v2grid}. The cells that completed are in Table~\ref{tab:v22legacy}; the
transcription floor under \texttt{noisy\_scan}, which is the comparator the v2.1
headline cell needs, and the hybrid agent were not run there. These rows inherit
every legacy-pool caveat in \S\ref{sec:limitations} and are exploratory.

\begin{table}[!htb]
\centering\footnotesize
\vtwotwotable{tables/v22_legacy_new_arms}{Mean expected loss per world for the arms added
in v2.2, on the legacy locked pool, by scan-noise regime, with paired contrasts against
\texttt{ocr\_rules\_book4} and \texttt{agent\_v2\_vlm+glm-flash}.}
\caption{Arms added in v2.2, measured on the legacy locked pool (exploratory).}
\label{tab:v22legacy}
\end{table}

\paragraph{The subtle-wording slice: H5 is supported as registered, and the margin is a date format.}
Section~\ref{sec:limitations} notes that the strong floor may overstate what
deterministic rules achieve when injected defects carry overt narration. The subtle
slice keeps the taxonomy and pricing and rewords the evidence of four classes so
that it avoids the keywords the committed floors match. The floors' collapse on the
reworded classes (Table~\ref{tab:v22subtle}) is therefore by construction, and a rule
written against the new wording recovers those defects: the oracle, which is such a
rule, recovers all \vttHfiveInjected{}. H5 asks what the text agent does there under
clean scans. The registered reading is \vttHfiveVerdict{}, reported as exploratory
because the companion cell sits on the pool whose oracle guard failed:
$\Delta=$\,\vttHfiveDelta{} BRL \vttHfiveCI{} over \vttHfiveN{} of
\vttHfiveExpected{} worlds (floor R\$\vttHfiveFloorEL{}, agent R\$\vttHfiveAgentEL{}
per world; Holm-adjusted $p=\vttHfiveHolm$), and on the regenerated pool with the
original wording the same contrast is \vttHfiveCompDelta{} \vttHfiveCompCI{} over
\vttHfiveCompN{} worlds, so both halves of the registered ``invert'' condition hold.
That result stands as registered. What it is evidence of is narrower than the
hypothesis's name (``subtle semantics invert the ranking'').

\emph{The margin is one class, and that class was not reworded semantically.} The
registration (\S9.1) separates the four reworded classes into three whose change is
semantic and one, \vttHfiveSurfaceClass{}, whose change is a date format: the
delivery stamp is printed as day, month abbreviation and year instead of with
slashes. It requires that class to be reported as its own row and kept out of any
semantic aggregate. Table~\ref{tab:v22h5classes} does so. The floor's stamp pattern
matches only the slashed form and finds \vttHfiveSurfaceFloorTP{} of
\vttHfiveSurfaceN{}; the agent finds \vttHfiveSurfaceAgentTP{} of
\vttHfiveSurfaceN{}. That class alone accounts for \vttHfiveSurfaceShare{} BRL per
world, which is \vttHfiveSurfacePct{}\% of the registered difference: more than all of
it, because the other classes taken together are against the agent. The agent did not get
better at dates: with the original wording it finds the same class
\vttHfiveCompSurfaceAgentTP{} of \vttHfiveCompSurfaceN{} and the floor
\vttHfiveCompSurfaceFloorTP{} of \vttHfiveCompSurfaceN{}. The ranking inverts because
the floor's pattern stops matching.

\emph{On the semantic classes the agent recovers less than half.} It finds
\vttHfiveSemTP{} of \vttHfiveSemN{} (floor: \vttHfiveSemFloorTP{}):
\vttHfiveRcptTP{} of \vttHfiveRcptN{} leisure receipts, which carry
R\$\vttHfiveRcptExposure{} of exposure per world, \vttHfiveDanfeTP{} of
\vttHfiveDanfeN{} non-creditable purchase descriptions (R\$\vttHfiveDanfeExposure{})
and \vttHfiveMixedTP{} of \vttHfiveMixedN{} private-expense narrations
(R\$\vttHfiveMixedExposure{}). Together they contribute \vttHfiveSemShare{} BRL per
world. The agent's pre-pass is \texttt{malha\_rules}, not the strong floor's rule
set, so on the \vttHfiveOtherN{} defects of classes that were not reworded it finds
\vttHfiveOtherAgentTP{} where the floor finds \vttHfiveOtherFloorTP{}, and it raises
\vttHfiveOtherAgentFP{} false positives there; those classes contribute
\vttHfiveOtherShare{}, more than the semantic classes give.

\emph{Without the date-format class the contrast is not in the agent's favour.}
Post hoc, with that class's defects and findings removed from both arms and
everything else scored as registered, $\Delta$ is \vttHfiveNoSurfDelta{}
\vttHfiveNoSurfCI{}: the floor is cheaper in \vttHfiveNoSurfFloorCheaper{} worlds,
the agent in \vttHfiveNoSurfAgentCheaper{}, \vttHfiveNoSurfTied{} tied. Across the six registered
price cells that reading is unresolved in \vttHfiveNoSurfGridUnresolved{} and
favours the floor in \vttHfiveNoSurfGridOpposite{} (at the R\$1{,}500 false-positive
price, neutral profile: \vttHfiveNoSurfHighFPDelta{} \vttHfiveNoSurfHighFPCI{}). The
same split by world: the \vttHfiveWithN{} worlds that hold a date-format defect give
\vttHfiveWithDelta{} \vttHfiveWithCI{}, the \vttHfiveWithoutN{} that do not give
\vttHfiveWithoutDelta{} \vttHfiveWithoutCI{}, with the floor cheaper in
\vttHfiveWithoutFloorCheaper{} of them (Table~\ref{tab:v22h5contrasts}). Restricted
to the three semantic classes alone the agent is ahead by \vttHfiveSemOnlyDelta{}
\vttHfiveSemOnlyCI{}, which is the recall above priced and is smaller than what it
loses on the classes that were not reworded. These slices were chosen after the
result was seen; by \S9.4 of the registration per-class contrasts on this slice are
descriptive.

\emph{False positives on benign texture} (\S9.3). The slice gives every receipt,
purchase DANFE, delivery stamp and a fixed number of ledger narrations a benign
wording of the same shape as a defect. Of the agent's \vttHfiveAgentFP{} false
positives, \vttHfiveTextureFP{} name a reworded class and cite a texture item of
that shape, and \vttHfiveTextureFPAny{} cite a texture item under any class; no
deterministic floor has one (Table~\ref{tab:v22h5texture}). The false positives
were not audited by hand.

The remaining qualifications are unchanged. The per-world split of the registered
contrast is \vttHfiveAgentCheaper{} worlds for the agent and \vttHfiveFloorCheaper{}
for the floor (sign test $p=\vttHfiveSignP$; median \vttHfiveMedian{}). The agent
has \vttHfiveAgentTP{} true positives (floor: \vttHfiveFloorTP{}) and
\vttHfiveAgentFP{} false positives (floor: \vttHfiveFloorFP{}); the registered
reading holds in \vttHfiveGridSupported{} of the \vttHfiveGridCells{} registered
price cells and is unresolved in the \texttt{conservative} profile at the R\$1{,}500
false-positive price. The claim licensed is about rules \emph{frozen before
the wording bank was written}; it is not that rules cannot handle such wording, and
for the class that carries the margin the missing rule is a second date pattern. A
floor with that pattern was not written or run, so the contrast without the class
is the nearest available reading of it, not a measurement of it. The agent is also
cheaper than every other deterministic floor measured in this cell
(Table~\ref{tab:v22floormatrixsubtle}, post hoc), for the same reason: none of them
parses the stamp. The agent arms were not run on the subtle slice under scan noise,
and the hybrid was not run on it at all, so this is one cell: text agent, clean
scans, 30 worlds, one model, one episode per world. In the registration's own
words for the alternative outcome, the measured agent mostly ``did not read the
semantics either''.

\begin{table}[!htb]
\centering\footnotesize
\vtwotwotable{tables/v22_h5_classes}{H5 by defect class: exposure, true and false positives of
both arms and each class's share of the registered difference.}
\caption{H5 by defect class (subtle slice, \texttt{clean}; strong floor against the
text agent). BRL per world over the 30 worlds. ``Recovered by agent'' is the
exposure the agent recovers over the floor, ``FP cost'' the difference in
false-positive cost, and the last column their sum; the rows add up to the
registered contrast. The date-format class is its own row and is not part of the
semantic total (registration \S9.1). Descriptive (\S9.4).}
\label{tab:v22h5classes}
\end{table}

\begin{table}[!htb]
\centering\footnotesize
\vtwotwotable{tables/v22_h5_contrasts}{The H5 contrast on class slices and world splits.}
\caption{The H5 contrast on class slices, by presence of a date-format defect, and
without the date-format class across the registered price cells. Only the first row
is a registered reading; every other row is post hoc.}
\label{tab:v22h5contrasts}
\end{table}

\begin{table}[!htb]
\centering\footnotesize
\vtwotwotable{tables/v22_subtle}{Mean expected loss per world and per-class recall on the
subtle slice for the deterministic floors and the agent arms, with paired
floor-minus-agent contrasts.}
\caption{The subtle-wording stress slice (exploratory): deterministic floors in every noise level, the text agent under \texttt{clean} only. Recall columns: the class whose rewording is a date format (date), the three semantically reworded classes (sem.), and the scan class whose wording is unchanged (orig.).}
\label{tab:v22subtle}
\end{table}

\paragraph{Model class and the ledger hypothesis: no evidence.} v2.1 measured
flash-class models only, and H2 of the second registration had no over-flagging arm
to act on (\S\ref{sec:limitations}). An exploratory lane outside the registration
started one arm on a different model and stopped below its own completion
rule; no SQL-versus-ledger contrast and no closed-lab model was run
(Table~\ref{tab:v22frontier}). No reading is drawn from it and both limitations stand.

\begin{table}[!htb]
\centering\footnotesize
\vtwotwotable{tables/v22_frontier}{Non-flash arm(s): mean expected loss, false positives
per episode, and the SQL-minus-ledger paired contrast; or an explicit ``not run'' row.}
\caption{The exploratory non-registered model lane: incomplete, no reading drawn.}
\label{tab:v22frontier}
\end{table}

\paragraph{The simplest reading, and the strongest objection to it.} On these worlds
the defects are written by a program, and a second program that reads the same words
finds them. Scan noise matters because it breaks the reading step; repair the reading
step, by deskewing or by one transcription call per page, and the rules are back at
their ceiling. A flash-class model in a 25-call loop reads the pages worse than the
same model asked to transcribe them, and what it adds on top of a rules pre-pass is
about one real catch per \vttHoneFPperTP{} false alarms. The objection is that the
agent side was tested with one inexpensive model, one prompt, one episode per world
and a prompt that makes rejected findings free, against rules that are the inverse of
the generator. All of that is true, and the H5 cell was built to show it. What that
cell shows instead is narrower: the agent is cheaper than every frozen floor there
because of a date format the floors' stamp pattern does not parse, and on the
classes whose meaning was reworded it finds \vttHfiveSemTP{} of \vttHfiveSemN{}
defects and is not cheaper once the date-format class is set aside
(\vttHfiveNoSurfDelta{} \vttHfiveNoSurfCI{}, post hoc). The objection therefore
stands untested rather than answered in the registered design: the semantic-only
slice is positive post hoc (\vttHfiveSemOnlyDelta{} \vttHfiveSemOnlyCI{}), but it
does not pay for the agent's losses on unreworded classes and false positives, and a
confirmatory test needs a slice whose margin cannot come from a surface format.

\paragraph{What v2.2 does not change.} The 8B evidence (\S\ref{sec:trainability}), the
absence of a human accountant baseline, the single task mode and the synthetic,
single-establishment scope are untouched by v2.2. Per-class recall, the sensitivity
grid, strict-citation rescoring and API cost for v2.2 are in Appendix~\ref{app:v22}.
\clearpage 
\fi

\section{Preliminary 8B trainability diagnostics}
\label{sec:trainability}

The 8B evidence in this version is diagnostic only. It is not a confirmatory
trainability result: v1 traces overlap locked evaluation seeds, v2 expert traces
were synthesized with oracle knowledge, the effective SFT window omitted the
flag/submit targets, historical eval rows used asymmetric tool availability, and
local GRPO is a bounded compute approximation rather than full GRPO. Within those
limits, corrected ledger-shaped GRPO is mechanically alive but behaviourally absent:
the locked-test policy emits zero ledger-receipted model-origin findings in 30/30
episodes, exactly matches SFT priced loss, and fails the pre-registered shape gate
(26 median tool calls versus the required $\leq 10$). Because the measurement is one
local training run and fails that gate, it is a negative pilot, not the H4 reversal
measurement. The detailed training trace, contamination analysis, and correction
history are quarantined to Appendix~\ref{app:trainability} and
\texttt{docs/TRAINING\_DIAGNOSTIC.md}; they should not be read as evidence of a
general 8B imitation ceiling.

\section{Limitations}
\label{sec:limitations}

The limitations are grouped by what they restrict, and each names the artifact or
decision entry that documents it. The paragraph on v2.2 is new in this revision; the
paragraphs on the legacy generator and the scan-inventory confound are updated for it.

\paragraph{v2.2 is exploratory.} The oracle guard failed on the
regenerated pool, so no claim on it is confirmatory, and that cannot be repaired
without a new pool under a new registration; completing the cells did not change
that. The registered agent cells are complete (\vttHoneN{} of \vttHoneExpected{}
worlds, and \vttHfiveN{} of \vttHfiveExpected{} for H5) but were run in two provider
usage windows about four hours apart, with
one episode per world, one flash-class model and one prompt; that prompt tells the
model a rejected finding is free. Two worlds entered after a retry that was not a
provider error. Without world \vttHtwoRetrySeed{} (pixel agent; the model's first
attempt ended on an invalid action) H2 is \vttHtwoNoRetryDelta{}
\vttHtwoNoRetryCI{} and H3 is \vttHthreeNoRetryDelta{} \vttHthreeNoRetryCI{} over
\vttHtwoNoRetryN{} worlds; without world \vttHoneRetrySeed{} (hybrid; an environment
import error before any model call) H1 is \vttHoneNoRetryDelta{}
\vttHoneNoRetryCI{} over \vttHoneNoRetryN{} worlds. No decision changes. On the
regenerated pool the pixel agent and the hybrid were run under
\texttt{noisy\_scan} only and the text agent under \texttt{clean} and
\texttt{noisy\_scan},
so the v2.1 clean-to-noisy degradation comparison can be re-checked for the text
agent alone. On the subtle slice only the text agent under \texttt{clean} was run.
The H5 reading rests on 30 worlds, on a false-positive count that was not audited by
hand, and on rules frozen before the subtle wording was written; its margin comes
from the date-format class, and the contrast without that class, the split by
world and the semantic-class slice are post hoc (D058). No floor with a second date
pattern was run. The H1 mean depends
on the assumed false-positive price, and the false positives were not audited by
hand. H2 compares two systems that differ in image resolution, call budget and
pre-pass rules as well as in having a loop. The transcription floor depends on a
provider model: its transcripts are provider output that cannot be regenerated. The
transcripts of the complete cells are committed with the repository (D055), so the
scoring of that floor can be rerun without a provider call, but the transcription
itself is a single run of a hosted model and is not reproducible. That floor reaching the perfect-text
ceiling also says the synthetic scan noise is easy for this model, and it need not
hold on real scans. The deskew recipe was chosen on development seeds and raises loss
on clean images. In the regenerated worlds the delivery-cutoff defect always sits on
one of the later delivery receipts by document number (found in an independent check
of the pools), a positional prior an agent could use without reading the scan;
whether any arm does was not tested. The
contrasts in Tables~\ref{tab:v22perworld} and~\ref{tab:v22replication} other than
H1--H4 are post hoc.

\paragraph{Sample, scope and variance.} The v1 and v2/v2.1 studies each evaluate 30 locked worlds; primary API cells use 3 episodes per arm while
scripted and local 8B cells use one deterministic/single training-run episode. v2.2 uses the pools in Table~\ref{tab:v22design} with one episode per world. Episode-level variance
is substantial, so paired-world bootstrap intervals are honest but wide. Only one task
mode (\texttt{auditoria}) is measured to completion. Generated companies are synthetic
\emph{lucro real} single-establishment entities covering a declared subset of SPED
records (26 ECD, 33 EFD-Contribui\c{c}\~oes; \texttt{docs/COVERAGE.md}); transfer to
real private filings is unmeasured. Within-world episode spread is no longer hidden by world averaging:
\texttt{analysis.json} now reports neutral/R\$250 mean, median, maximum episode-level
SD and mean episode range for every arm$\times$noise cell.

\paragraph{Pricing and cost.} Although v2 evaluates the full $3 \times 2$
sensitivity grid ($P(\text{autua\c{c}\~ao}) \in \{0.05, 0.15, 0.40\} \times
c_{\text{fp}} \in \{\text{R\$}250, \text{R\$}1{,}500\}$), API token costs are
reported separately rather than added to statutory exposure
(Table~\ref{tab:v2costs}: mean \$0.028--\$0.031/episode for
\texttt{deepseek-v4.1-flash}, \$0.036--\$0.041 for \texttt{glm-5.3-flash}, and
\$0.252 for \texttt{gemini-3.8-flash} where usage capture exists).

\paragraph{Post-hoc status of v2.1.} The v2.1
capability separation (D033) is a post-audit amended analysis after
fixing three v2.0 harness defects, with all frozen v2.0 rows preserved alongside it.
Because the v2.1 prompt rebalance and the \texttt{book4} floor rules were designed after
inspecting failures on the same locked pool, those v2.1 separations are exploratory
amended analyses rather than independent confirmatory tests.

\paragraph{Leakage and oracle blinding.} The pre-D042 v1 filesystem also exposed \texttt{world.json}; because that file
contained defect-class metadata, v1 and \texttt{agent\_v1} measurements collected before
the correction are exploratory. Audit of v2 \texttt{agent\_v2\_*} prompts and
trajectories found no reads of \texttt{world.json}, but future confirmatory contrasts
must use the patched filesystem. Separately, the oracle gate is not evidence-free answer-key
blinding for every class: injected defect metadata can target which artifact evidence is
recomputed, especially for substantive classes, although agent output is never accepted as
truth.

\paragraph{The strong floor.} The strongest-floor correction further limits the capability claim: agents beat the
original rules engine on the mean on both pools, and the naive OCR floor only in
legacy-pool cells that did not replicate (\S\ref{sec:v22}), but not
\texttt{ocr\_rules\_book4}. The floor's ceiling is partly circular: its rules and the
defect injector share a keyword list, and on the subtle slice, where the wording
avoids those keywords, the same rules recover no instance of the four reworded classes
(Table~\ref{tab:v22subtle}); for one of the four the change is a date format, not a
meaning (Table~\ref{tab:v22h5classes}). Moreover, the four added floor rules are intentionally
artifact-level auditor heuristics over the current generator; to the extent injected
defects carry overt narration or structural signatures, the floor may overstate what a
real-world deterministic system would achieve.

\paragraph{Legacy generator and renderer.} D046 corrects several SPED
renderer semantics only for newly generated worlds: the legacy locked pool
(\S\ref{sec:results}) was generated before that correction, uses an annual single-file
EFD-Contribui\c{c}\~oes representation, and remains a synthetic statutory approximation
rather than a validated representation of production filings. The v2.2 pools are
generated with the corrected renderer and carry their own \texttt{gen\_version}
(Table~\ref{tab:v22design}); because v2.2 is exploratory, the corrected
renderer is in use but not yet backed by a confirmatory measurement.

\paragraph{Scan-inventory confound.} On the legacy pool the scan generator appended defect-carrying scans to a baseline scan plan,
so scan inventory shape could itself cue the presence of a defect independent of pixel
evidence. The locked-pool measurement artifact (\texttt{results/v2/scan\_inventory\_shape.json})
quantifies the cue: worlds with zero, one, or two scan-class defects average 19.27,
20.0, and 22.0 visible scan PNGs respectively. We found no evidence that the audited
agents explicitly exploited this cue. The regenerated pool removes it
(Table~\ref{tab:v22inventory}), and the two v2.1 pixel-vision contrasts did not
replicate there (Table~\ref{tab:v22replication}); whether the cue, the thin legacy
scan slice (3--6 instanced worlds per scan class, \S\ref{sec:results}) or chance
produced the legacy result cannot be told apart from these data.

\paragraph{Untested hypotheses and missing comparators.} On H2, neither flash arm is an over-flagger, so fail-closed FP suppression remains
untested on high-FP models. The v2.1 floor contrast is robust but not universal: under conservative
enforcement with R\$1{,}500 false-positive rework, mean hybrid-agent advantages remain
positive but paired bootstrap intervals cross zero. All measured v2 API arms are
flash-class models with the family substitutions recorded in D030 and D037; no
non-flash frontier model is measured on the v2 surface, and the incomplete exploratory
lane in Table~\ref{tab:v22frontier} does not change that. On H4, the corrected ledger-shaped
8B MLX GRPO measurement is a single local training run with the compute simplifications
and fresh-optimizer crash resume documented in D039--D040, not the pre-registered
three-training-seed confirmation. Its negative result is nevertheless behaviorally
strong: zero model-origin findings in 30/30 locked episodes, exact SFT-equivalent
priced loss, and failure of the shape gate. Finally, there is no human accountant
baseline: what a practitioner costs on these worlds remains the most valuable
unmeasured comparator.

\section{Conclusion}
\label{sec:conclusion}

Priced measurement changed the verdict three times, each time against the agent. In
v1, standalone frontier agents beat trivial floors yet failed to beat the
deterministic rules engine; those measurements are exploratory because the
pre-correction filesystem exposed defect metadata. In v2.1, moving the deterministic
layer into a mandatory pre-pass outside the agent's budget, replacing pagination with
a read-only SQL index and fail-closed Trust Ledger, and fixing the D033 harness
defects made the post-hoc hybrid agents lower-loss than the original
\texttt{malha\_rules} floor in exploratory contrasts, and that sign held on the
regenerated pool. Then four visible-artifact auditor rules added to the OCR floor
priced lower than the best agent on every noise regime of the legacy pool.

The scan layer is where this revision takes most back. On the legacy pool, 20 lines
of Tesseract regex degraded +77.4\% from clean to noisy scans while the pixel-vision
agent degraded +5.0\% and separated ahead of that naive floor. On the regenerated
pool, with a defect-independent scan inventory, the pixel agent is not ahead of the
naive floor (\vttNaiveVlmDelta{} \vttNaiveVlmCI{}) and not ahead of the same model
reading OCR text (\vttHthreeDelta{} \vttHthreeCI{}); the degradation comparison can
be re-checked there for the text agent only, because the pixel agent was not run on
clean scans. We retract the v2.1
pixel-vision readings as general claims and keep the numbers as legacy-pool
measurements. What the regenerated pool does support, as exploratory evidence, is
about reading rather than auditing: deskewing before OCR recovers part of what scan
noise costs the deterministic rules, one transcription call per page from the vision
model returns them to their clean-scan loss, and both pipelines are lower-loss than
the rules-plus-agent hybrid.
That hybrid is cheaper than the raw-OCR strong floor on the mean and dearer in
\vttHoneFloorCheaper{} of \vttHoneNonTied{} non-tied worlds, a result that turns on
the assumed price of a false positive.

None of this is confirmatory. The v2.2 oracle guard failed and it tested one
flash-class model with one prompt. The one hypothesis about evidence the rules were
not written against is supported as registered (\vttHfiveDelta{} BRL
\vttHfiveCI{} on \vttHfiveN{} worlds, at \vttHfiveAgentFP{} false positives to the
floor's \vttHfiveFloorFP{}), and it does not show what its name suggests. The whole
margin is a delivery-stamp date format that the floor's pattern does not parse; on
the semantically reworded classes the agent finds \vttHfiveSemTP{} of
\vttHfiveSemN{} defects, and without the date-format class the contrast is
\vttHfiveNoSurfDelta{} \vttHfiveNoSurfCI{} (post hoc). Whether an agent earns its
cost where meaning, not format, leaves the rules' vocabulary is still open, and the
rules' ceiling on these worlds is partly the generator agreeing with itself. The
strongest defensible claim is therefore not that agents surpass deterministic
automation, and not that they cannot. It is that statutory priced loss makes the
comparison measurable, and that in every cell where an agent was measured the
cheapest arm is a pipeline with no agent loop (the same rules over OCR text or over
one transcription call per page), except the H5 cell, where the agent's lead over
rules frozen beforehand is a date format. Current 8B SFT/GRPO evidence is diagnostic only. The operational
takeaway for regulated accounting is: \textbf{never replace deterministic verification
with an LLM, repair the reading step before adding an auditor loop when physical
documents degrade, never evaluate an auditor without statutory money and the price of
a false alarm on the axes, and never claim agent capability against a floor you have
not maximized.}

\section*{Disclosures and availability}
\label{sec:availability}

The code, the analysis scripts and the text were written with an AI coding assistant,
under the authors' direction and review. Caio Theodoro works on a commercial AI gateway;
nothing from that product is used, every component is open source and all data is
synthetic. The corrected public artifact package --- analysis tables and JSON, this
PDF, the slim source/analysis snapshot, compressed locked-pool and trajectory archives,
and SHA-256 checksums --- is under \texttt{corrected/} in the public Hugging Face dataset
\url{https://huggingface.co/datasets/caiotheodoro/malha-benchmark/tree/main/corrected}.
The broader Hugging Face collection \texttt{caiotheodoro/malha} contains the benchmark,
traces, and diagnostic adapter repositories. That package predates v2.2: the v2.2
episodes, analysis files and generated tables are under \texttt{results/v22/} in the
development repository and are not yet part of it. The paper is CC BY 4.0; the code is Apache-2.0.

\bibliographystyle{plainnat}
{\footnotesize\setlength{\bibsep}{0pt}\bibliography{refs}}

\appendix

\section{Registrations and deviations}
\label{app:registration}

\paragraph{Tags in this repository.} The repository this revision is built from
carries one registration tag, \texttt{prereg-v3}. The tags \texttt{prereg-v1} and
\texttt{prereg-v2} named below were made in an earlier development repository and
are not present in this one; the three frozen registration texts are. A reader of
this tree can therefore check the v2.2 freeze against a tag and the two earlier
freezes only against the file contents and the decision log.

\paragraph{The registration.} \texttt{PRE-REGISTRATION.md}, frozen at tag
\texttt{prereg-v1} before the first measured run, fixes the hypotheses H1--H3 and the
decision rules, including the honest-negative branch: if agents do not separate from
the deterministic engine, that is the result and it is published. Both branches of
each hypothesis were written before measurement.

\paragraph{v2 registration provenance.} A lightweight tag \texttt{prereg-v2}
exists at \texttt{9de307d0} in the earlier development repository (neither the tag
nor that commit is in the repository this revision is built from) and predates the v2
measurement rows. Because the development repository is not yet public, that tag is not independently
inspectable by arXiv readers; the frozen registration text is included in the public
source snapshot, but the snapshot timestamp does not retroactively prove the tag date.
v2 pre-registration provenance is therefore stronger than pure post hoc analysis inside
the project history but weaker than a public signed/annotated or immutable-registration
artifact. The v2.1 strong-floor correction (D042--D044) is explicitly post hoc and
registered as such rather than retroactively treated as part of the original H1/H3
decision rule.

\paragraph{Post-freeze decisions (v1).} Every post-freeze interpretation is a dated entry in
\texttt{docs/DECISIONS.md}. The v1 entries the paper depends on: D007 (citation-anchored
finding matching; blanket findings score 0.0 recall); D008 (CNPJ wording tightened to
the declared \texttt{99.xxx.xxx} synthetic base convention); D012 (the world, not the
episode, is the unit of randomness; the bimodal whiff pattern first observed in the
flash arm); D014 (C3 training path amended from Tinker to free-tier Kaggle T4 on
\texttt{Qwen3-8B} after Tinker billing block); D016 (the hydrogen $n{=}30$ verdict; the
honest-negative branch stands because superiority over the engine is not established);
D017--D018 (SFT-v2 trained clean, final loss 0.390, and fails behaviourally with the
mechanism observed in the probe); D019--D020 (the GRPO smoke and the zero-variance dead
end, written as a pre-registered reversal once pattern-visible seeds confirmed it);
D021 (deepseek $n{=}30$ separated worse than \texttt{malha\_rules}); D023 (the
class-conditional slice, computed after the first draft because the comparison demanded
it); D024 (the loss-profile sensitivity scan, with the caveat that trajectory
re-scoring diverges slightly from the committed pipeline's point values, so it carries
the regime shape while Table~\ref{tab:headline} carries the headline estimates).

\paragraph{v2.2 registration provenance.} \texttt{PRE-REGISTRATION-V3.md} is frozen at
tag \texttt{prereg-v3}; the commit, tag time and file hash are printed with
Table~\ref{tab:v22design}. The tag was made before the regenerated pools were built,
in the private development repository, and had not been pushed when it was made, so
like \texttt{prereg-v2} it is not an immutable public timestamp. Two decision entries
follow the freeze: D053 (generator 2.2 and the integration choices made at the
freeze) and D054 (the failed oracle guard, the agent cells as they stood when the provider limit
first stopped them, the ruling
that a world lost to a model-emitted invalid action stays lost, and the registered
analysis script as built, which was edited after the freeze to apply the registered
cell and multiplicity rules). Two further entries are bookkeeping: D055 (provider
account identifiers redacted from stored error strings; transcripts of the complete
transcription cells committed) and D056 (the post hoc analysis added after the
verdicts were first read, and the ruling that the regenerated pool stays exploratory:
no amended guard is applied to it and no new registration is opened in this
revision). D057 records the completed measurement: the resumed cells, the two
retried worlds and their with/without readings, H5 moving from not run to
\vttHfiveVerdict{} (exploratory), and the committed scan inventory that lets the
analysis be rebuilt without regenerating the pools. D058 corrects how H5 is
reported: the per-class rows, the separate row for the date-format class and the
count of false positives on benign texture that \S8, \S9.1 and \S9.3 of the
registration require were missing from the first complete build and are added, with
post hoc contrasts on class slices, and the prose no longer describes the H5 margin
as a result about wording or meaning. The per-world counts, the replication table in \S\ref{sec:v22} and
Tables~\ref{tab:v22floormatrix}--\ref{tab:v22costdetail} are not part of the
registered analysis.

\paragraph{Post-freeze decisions (v2 and v2.1).} The v2-era entries the paper depends
on, by topic. \emph{Registration and surface:} D025 (v2 generator corrections and
pre-registration freeze); D026 (surface compression; the horizon gate); D027 (citation
schema and rejection messages); D028 (scorer false-positive semantics, adjacency backing
for absence evidence, scripted-expert trace synthesis); D029 (API cost capture).
\emph{Arms and models:} D030 (the deepseek arm is \texttt{deepseek-v4.1-flash});
D037 (flash-class model-substitution confounds in the v2.1 API arms); D041 and D050
(the same-model \texttt{agent\_v1} contrast and its post-D042 leak-free rerun).
\emph{Harness audit and interpretation:} D033 (the v2.1 harness fixes and the
pixel-vision arm); D042 (anti-leakage correction and the book-side deterministic floor);
D043 (corrected interpretation: the strong floor is ahead of every measured agent); D044
(strict-citation sensitivity); D047 (within-world spread and incomplete-cell coverage);
D049 (scan-inventory confound and citation-stuffing control); D052 (the completed
same-model \texttt{fixed+glm-flash} control, the scan-inventory artifact, and the
registration-provenance correction). \emph{Generator:} D046 (forward SPED renderer
correction; the measured pool predates it); D051 (generator version bump for newly
generated worlds). \emph{8B training:} D031--D032, D034--D036, D038--D040 and D045
(Appendix~\ref{app:trainability}).

\paragraph{Deviations (v1).} The \texttt{pro} arm was planned at the same coverage as the
other Gemini arms and was abandoned after persistent rate limiting ($n{=}4$ worlds);
its evidence is marked partial wherever used. Two hydrogen batch processes were killed
by the host mid-run; the affected seeds were re-run in full, and only complete episodes
enter the tables. The deepseek arm ran under a windowed quota across several days; the
windows do not enter the loss model. None of these affects a frozen file; all are in
the decision log with dates.

\paragraph{Codenames.} The v1 arm labels \texttt{gemini-hydrogen} and
\texttt{deepseek-v4-pro} are codenames for dated vendor model versions; the v2 arm
labels (\texttt{deepseek-flash}, \texttt{glm-flash}, \texttt{gemini-flash}) are distinct
arms, not renamings of the v1 arms (D030). The measurement dates and exact version
strings are in the repository's decision log; the models are named openly at
publication.

\section{8B training diagnostic detail}
\label{app:trainability}

This appendix is ordered by relevance rather than chronology: first the v2 pipeline
measured on the locked v2 pool (SFT, then corrected ledger-shaped GRPO), then the
archived v1 free-tier pipeline that preceded it. Both are diagnostic
(\S\ref{sec:trainability}).

\paragraph{Behavioral failure in the diagnostic SFT checkpoint.} Stage-1 SFT on verified
expert traces produces format without judgment, in both directions the
distribution permits. The first checkpoint (shallow expert traces, mean five
calls) learned \emph{instant-submit}: two-call episodes whose measured true
positives were entirely the deterministic baseline's, an auditor that bills
for the pre-pass and goes home. Regenerating the traces with mandatory
investigation depth (D031) and retraining produced the mirror failure:
episodes that consume the full 25-call budget in well-formed, correctly-cited
investigation --- survey, scan inspection, cross-checks --- and never flag
anything, looping \texttt{inspect\_scan} over sequential receipt ids with
verbatim narration. Both checkpoints pass naive format checks; neither audits.
The behavioral gate the pre-registration requires exists precisely because
these are distinguishable only behaviorally: the strengthened gate (flags on
\emph{each} probe world) fails the second checkpoint 0/3. The locked-seed
measurement of this checkpoint is internally consistent but diagnostic rather than
confirmatory for 8B trainability in general: across all thirty test worlds it emits \emph{zero} model-origin findings in 25-call
episodes whose every call is well-formed (mean 24.9/25) --- every true
positive in its scored episodes is the deterministic pre-pass's, and its mean
priced loss (R\$19{,}645) is higher than \texttt{malha\_rules} (R\$15{,}719)
because \texttt{AuditoriaEnvV2} routes \texttt{sequence\_gap} into open
warnings requiring adjudication, which the zero-action SFT policy never
performs. Format-perfect, judgment-absent in this diagnostic training setup. Because later
inspection found truncation/provenance defects in the training data and harness, this
must not be generalized into an imitation-ceiling claim; it shows only that the
measured pipeline failed behaviorally despite a low loss curve (0.059 at convergence). Subsequent Stage-2 GRPO smoke runs from both SFT and base
initialisations under unshaped terminal reward also recorded
\texttt{reward\_std} $= 0.0$ on 3/3 iterations (D032); the D033 audit
further identified that \texttt{training/mlx\_grpo.py} had omitted both
\texttt{tools=TOOLS\_V2} chat-template injection and the programmatic
ledger-shaped warm-up reward specified in \texttt{V2\_PLAN.md}~\S5.3.

\paragraph{Corrected shaped GRPO: mechanically alive, behaviourally absent.}
The corrected v2.1 MLX implementation completed all 16 bounded iterations after
the template-prefill, backward-kernel, reward-scaling, invalid-tool, and checkpoint
repairs (D036--D040). It is therefore no longer a zero-gradient scaffold: nonzero
within-group reward variance occurred on multiple training worlds, most strongly on
\texttt{seed306}, and the final iterations used the corrected baseline-aware,
receita-scaled ledger reward. The locked-test result is nevertheless negative.
Across seeds 501--530, \texttt{qwen3-8b-v2-grpo} emits \emph{zero} ledger-receipted
model-origin findings in 30/30 episodes and zero rejected flags; its priced loss is
byte-for-byte identical to SFT on every world (mean R\$19{,}645,
median R\$13{,}215). It does not beat the untrained base model
(R\$17{,}787 mean; paired GRPO-minus-base point estimate +R\$1{,}857, interval
crossing zero) and is decisively worse than \texttt{malha\_rules}
(R\$15{,}719 mean; rules advantage R\$3{,}926 [+772, +8{,}064]). It also fails the
pre-registered shape gate before any positive interpretation is permitted: median
episode length is 26 tool calls against the required $\leq 10$. Three episodes
collapse to one or zero well-formed calls, but the dominant failure is not syntax;
it is a policy that investigates without ever attempting a ledger-verified flag.
Because this is one local training run rather than the pre-registered three
independent training seeds, it is a strong negative pilot rather than the full H4
reversal-clause measurement; because the shape gate fails, it cannot be a positive
H4 result under any training-seed interpretation.

\begin{figure}[!htb]
\centering
\includegraphics[width=\textwidth]{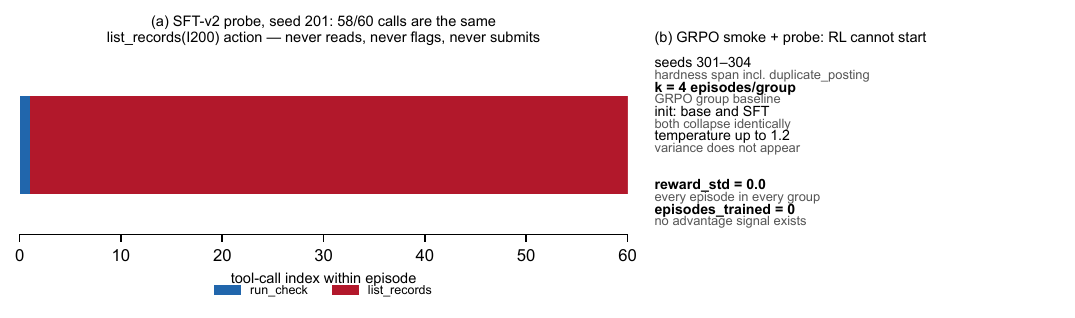}
\caption{The 8B collapse. (a) The probe episode: parseable syntax throughout, 58 of 60
calls the identical modal action, no engagement with documents. (b) The GRPO probe
configuration: across 4 seeds spanning the taxonomy's hardness range, both
initialisations, temperatures to 1.2, within-group outcome variance is exactly zero.}
\label{fig:collapse}
\end{figure}

\paragraph{The archived v1 training pipeline also fails behaviourally.} The earlier
free-tier training path was executed end to end on \texttt{unsloth/Qwen3-8B-bnb-4bit}
with LoRA ($r{=}16$, $\alpha{=}32$, all attention and MLP projections) on a free Kaggle
Tesla T4 and measured at each gate (decision trail D014, D017--D020). 80 verified traces
(5 rules-gold, 75 perfect frontier episodes from the C2 pool) were curated with
queue-based tool-call pairing; SFT-v2 converged cleanly on a free T4 to final training
loss 0.390. Protocol-matched behavioural evaluation on locked seeds 201--205 ($k{=}1$,
\texttt{MAX\_CTX}$=6144$, 60-call budget, temperature $0.0$) is reported in
Table~\ref{tab:qwen_eval} alongside the base model (\texttt{qwen\_eval\_base\_rows.json}).
SFT-v2 flags \emph{zero} findings on every seed --- mean loss 17{,}184.79 BRL,
indistinguishable from \texttt{file\_as\_is} and 5{,}775 BRL \emph{worse} than the
untrained base model (11{,}409.76 BRL). The probe shows why: SFT-v2 emits perfectly
parseable tool-call syntax, then loops the modal opening action --- 58 of 60 calls are
the identical \texttt{list\_records} call --- and never reads, never flags, never
submits (Figure~\ref{fig:collapse}a). Imitation learned the shape of the opening moves
and none of the investigation. The base model is erratic but non-degenerate: it catches
one defect on seed 201 (3 findings, 2 FPs) and spams 53 findings (1 TP, 52 FPs) on
seed 202, so behavioural and outcome variance exist before SFT.

\begin{table}[!htb]
\centering\small
\begin{tabular}{lcccc}
\toprule
seed & base findings (TP/FP) & base loss (BRL) & SFT-v2 findings & SFT-v2 loss (BRL) \\
\midrule
201 & 3 (1/2) & 14{,}322.70 & 0 & 55{,}820.19 \\
202 & 53 (1/52) & 15{,}084.48 & 0 & 2{,}462.10 \\
203 & 0 (0/0) & 25{,}008.96 & 0 & 25{,}008.96 \\
204 & 0 (0/0) & 1{,}766.89 & 0 & 1{,}766.89 \\
205 & 0 (0/0) & 865.79 & 0 & 865.79 \\
\midrule
mean & 11.2 (0.4/10.8) & 11{,}409.76 & 0.0 & 17{,}184.79 \\
\bottomrule
\end{tabular}

\caption{Qwen3-8B base vs.\ SFT-v2 behavioural evaluation on locked seeds 201--205
($k{=}1$, 60-call budget, temperature $0.0$). SFT-v2 emits zero findings on every seed
(loss equals unaudited exposure), worsening mean loss by +5{,}775 BRL relative to the
erratic base model while eliminating all behavioural variance.}
\label{tab:qwen_eval}
\end{table}

\paragraph{Reinforcement cannot start (archived v1 pipeline).} Group-relative policy optimisation requires
outcome variance within a group: if all $k$ episodes of a world earn the same reward
($R = -\mathcal{L}$), every advantage is zero and no gradient exists. We ran
group-baseline REINFORCE (\texttt{MAX\_CTX}$=2048$, 40-call budget, $k{=}4$,
$\text{lr}=10^{-5}$, $\sim$2{,}700--4{,}300\,s per iteration on T4) across training
seeds 301--304 --- spanning the taxonomy's hardness range, including the
pattern-visible \texttt{duplicate\_posting} --- from both SFT-v2 and base
initialisations at temperatures up to $1.2$. On every seed, losses are byte-identical
across all $k{=}4$ episodes, within-group reward standard deviation is exactly zero,
and episodes trained is zero (Figure~\ref{fig:collapse}b).

The iron law of agentic RL is usually stated as: reinforcement needs a seed policy with
non-trivial solve rate. The inversion measured here is sharper. Under this free-tier
4-bit LoRA setup and sparse terminal reward, imitation destroyed the outcome variance
the base model had, collapsing erratic-but-varied behaviour into a single degenerate
fixed point and removing the signal reinforcement needed; even base initialisation
remained trapped in zero-variance rollouts on the probe seeds. Whether this collapse
is fundamental at 8B scale or an artefact of context truncation, sparse terminal
reward, or frontier-teacher traces unlearnably far from the student's policy is open;
the traces and the failed adapter are public so both hypotheses can be tested.

\section{Taxonomy and environment specification}
\label{app:spec}

Table~\ref{tab:taxonomy} lists the frozen defect classes: the 15 v1 classes
implemented in \texttt{src/malha/forge/defects.py} and the four scan-substantive classes
added in v2 (\S\ref{sec:envv2}). Layer and detectability are taken from
\texttt{taxonomy.json}; the pricing basis is the one the scorer applies,
\texttt{CLASS\_PRICING} in \texttt{src/malha/eval/loss.py}. Every world in the 30-world
v1 C2 pool (seeds 201--230) injects exactly one substantive class (middle block) and one
engine-detectable class (top block) in rotation; the scan block exists only in v2
worlds.

\begin{table}[!htb]
\centering\footnotesize
\begin{tabular}{llll}
\toprule
defect class & layer & detectable\_by & pricing basis \\
\midrule
\texttt{layout\_violation} & syntactic & L1 & presentation \\
\texttt{control\_total\_forgery} & syntactic & L1 & presentation \\
\texttt{unbalanced\_entry} & intra\_ledger & L2 & presentation \\
\texttt{coa\_referencial\_gap} & intra\_filing & L2 & presentation \\
\texttt{j100\_vs\_i155\_break} & intra\_filing & L3 & operation \\
\texttt{duplicate\_posting} & intra\_ledger & L3 & operation \\
\texttt{unrecorded\_nfe} & cross\_filing & L3 & operation + autua\c{c}\~ao \\
\texttt{nfe\_value\_mismatch} & cross\_filing & L3 & operation \\
\texttt{apuracao\_vs\_dre} & cross\_filing & L3 & operation \\
\texttt{participant\_master\_break} & cross\_filing & L3 & presentation \\
\texttt{sequence\_gap} & intra\_filing & L3\_assist & presentation \\
\midrule
\texttt{cutoff\_violation} & substantive & L3\_assist & substantive \\
\texttt{pis\_cofins\_ineligible\_credit} & substantive & agent\_judgment & substantive \\
\texttt{transitoria\_parking} & substantive & agent\_judgment & substantive \\
\texttt{expense\_personal\_mixed} & substantive & agent\_judgment & substantive \\
\midrule
\texttt{scan\_personal\_expense} & substantive\_scan & agent\_judgment & substantive \\
\texttt{scan\_cutoff\_delivery} & substantive\_scan & agent\_judgment & substantive \\
\texttt{scan\_value\_alteration} & substantive\_scan & agent\_judgment & operation + autua\c{c}\~ao \\
\texttt{scan\_ineligible\_credit\_desc} & substantive\_scan & agent\_judgment & substantive \\
\bottomrule
\end{tabular}

\caption{The frozen defect taxonomy: 15 classes in v1, 19 in v2. Top block: 11
engine-detectable classes (L1 layout, L2 invariants, L3 cross-filing rules, L3-assisted
sequence check). Middle block: 4 substantive classes excluded from \texttt{malha\_rules}
by design. Bottom block: the 4 scan-substantive classes added in v2, visible only in the
scan layer. Pricing bases map to Equation~\ref{eq:loss} and Lei 8.218/1991, art.~12.}
\label{tab:taxonomy}
\end{table}

\ifvtwotwo
\clearpage
\section{v2.2 generated detail}
\label{app:v22}

Every table in this appendix is generated; see \S\ref{sec:v22} for how to read them.
All of it is exploratory; agent rows exist under \texttt{noisy\_scan}, and under
\texttt{clean} for the text agent only.

\begin{table}[!htb]
\centering\footnotesize
\vtwotwotable{tables/v22_scan_noise}{Per-class recall on the four scan-substantive
classes by scan-noise regime and arm, with instance counts.}
\caption{v2.2 scan-class recall on the regenerated pool.}
\label{tab:v22scan}
\end{table}

\begin{table}[!htb]
\centering\footnotesize
\vtwotwotable{tables/v22_sensitivity}{Floor-minus-agent paired deltas across enforcement
probability and false-positive rework cost, separated cells marked.}
\caption{v2.2 sensitivity grid.}
\label{tab:v22sensitivity}
\end{table}

\begin{table}[!htb]
\centering\footnotesize
\vtwotwotable{tables/v22_strict_citation}{Mean expected loss under the primary scorer,
exact-citation strict rescoring and one- and three-citation caps, per arm and noise
regime.}
\caption{v2.2 strict-citation and citation-cap sensitivity.}
\label{tab:v22strict}
\end{table}

\begin{table}[!htb]
\centering\footnotesize
\vtwotwotable{tables/v22_api_costs}{Episodes with usage capture, prompt and output tokens,
total and mean cost per episode, per arm and noise regime, with the price list used.}
\caption{v2.2 measured API-token cost.}
\label{tab:v22costs}
\end{table}

\begin{sidewaystable}[p]
\centering\footnotesize
\vtwotwowidetable{tables/v22_floor_matrix}{Each agent arm against every deterministic floor
on the regenerated pool under noisy scans: paired delta, interval and per-world
counts.}
\caption{Post hoc: every floor against every agent arm on the regenerated pool,
\texttt{noisy\_scan}. Positive means the agent is cheaper; bold means the interval
excludes zero. Each cell uses the worlds on which both arms have an episode.}
\label{tab:v22floormatrix}
\end{sidewaystable}

\begin{table}[!htb]
\centering\footnotesize
\vtwotwotable{tables/v22_completeness}{Worlds done over planned for every agent cell and
every incomplete cell, with the reason and the missing seeds.}
\caption{Completeness of the agent cells and of the transcription caches, with the
reason each world is missing.}
\label{tab:v22completeness}
\end{table}

\begin{table}[!htb]
\centering\footnotesize
\vtwotwotable{tables/v22_floor_matrix_subtle}{Every deterministic floor against the text
agent on the subtle slice under clean scans: paired contrast and per-world split.}
\caption{Subtle slice, \texttt{clean}: every floor against the text agent (post hoc,
exploratory). The floors' keyword lists were frozen before the subtle wording was
written.}
\label{tab:v22floormatrixsubtle}
\end{table}

\begin{table}[!htb]
\centering\footnotesize
\vtwotwotable{tables/v22_h5_texture_fp}{False positives on benign texture per arm and noise
level on the subtle slice.}
\caption{Subtle slice: false positives on benign texture, per arm and noise level
(registration \S9.3). ``Same class'': a false positive of the reworded class a
texture item imitates, citing that item, by kind of item; ``any class'': any false
positive citing a texture item. Identifying texture by the cited item is a post hoc
definition.}
\label{tab:v22h5texture}
\end{table}

\begin{sidewaystable}[p]
\centering\footnotesize
\vtwotwowidetable{tables/v22_cost_detail}{List-price cost, tokens, requests and wall time per
episode for each agent arm, beside what the arm adds to three floors.}
\caption{Post hoc: measured cost per world beside what each agent arm adds to three
floors (true and false positives, and the paired delta). Token counts times list
price on a subscription key, not an invoice.}
\label{tab:v22costdetail}
\end{sidewaystable}
\fi

\end{document}